\documentclass{article}  
\usepackage{cite}
\usepackage{calc}
\usepackage{graphicx}
          \begin{document} 

		\title{An information theoretic analysis of decision in computer chess }   
		\author{Alexandru Godescu, ETH Zurich}
		\maketitle

	\newtheorem{axiom}{Axiom}
	\newtheorem{theorem}{Theorem}
	\newtheorem{definition}{Definition}	
	\newtheorem{proposition}{Proposition}
	\newtheorem{interpretation}{Interpretation}
	\newtheorem{analysis}{Analysis}
	\newtheorem{search experiment}{Search Experiment}
	\newtheorem{assumption}{Assumption}
	\newtheorem{principle}{Principle}
	\newtheorem{conjecture}{Conjecture}
	\newtheorem{positional analysis}{positional analysis}
	\newtheorem{positional judgment }{positional judgment}
	\newtheorem{experimental evidence}{experimental evidence}

	\begin{abstract}

	The basis of the method proposed in this article is the idea that information is one of the most important factors in strategic decisions, including decisions in
	computer chess and other strategy games. The model proposed in this article and the algorithm described are based on the idea of a information theoretic basis
	of decision in strategy games . The model generalizes and provides a mathematical justification for one of the 
	most popular search algorithms used in leading computer chess programs, the fractional ply scheme. However, despite its success in leading computer chess 				applications,  until now few has been published about this method. The article creates a fundamental basis for this method
	in the axioms of information theory, then derives the principles used in programming the search and describes mathematically the form of the 				coefficients. One of the most important parameters of the fractional ply search is derived from fundamental principles. Until now
	this coefficient has been usually handcrafted or determined from intuitive elements or data mining. There is a deep, information theoretical justification for such a parameter. In one way the method proposed is a generalization of previous methods. More important, it shows why the fractional depth ply scheme is so powerful. It is because the algorithm navigates along the lines where the  highest information gain is possible.   A working and original implementation has been written and tested for this algorithm and is provided in the appendix. The article is essentially self-contained and gives proper background knowledge and references. The assumptions are intuitive and in the direction expected and described intuitively by great champions of chess.

	\end{abstract}

	\newpage

	\section{Introduction}

	\subsection{ Motivation}

	Chess and other strategy games represent models of decision which can be formalized as computation problems having many similarities with important problems in computer science.
It has been proven that chess is an EXPTIME-COMPLET problem ~\cite{FL}, therefore it can be transformed in polynomial time in any problem belonging to the same class of complexity. Most of the methods used to program chess refer to the 8x8 case and therefore 
are less general. Such methods are not connected in their present form to the more general problems of complexity theory. A 
bridge may be constructed by generalizing the exploration and decision methods in computer chess. This is an important reason 
for seeking a more general form of these methods. In this regard a mathematical interpretation and description of information in the context of chess and computer chess may
be a condition. A second reason has to do with the gap in scientific publications about the fraction ply methods. As Hans Berliner pointed out about the scheme of ''partial depths'', ''...the success of these micros (micro-processor based programs) attests to the efficacy of the procedure. Unfortunately, little has been published on this''. A mathematical model of chess has been an interest of many famous scientist such as Norbert Wiener, John Von Neumann,
Claude Shannon, Allan Turing, Richard Bellman and many others. The first program has been developed by the scientist from Los Alamos National laboratory, the same laboratory that developed the first nuclear weapons. The first world champion program has been developed by the scientists form a physics institute in the former Soviet Union.
It has been speculated that chess may play a role in the development of artificial intelligence and certainly the alpha-beta method, used now in all adversarial games 
has been developed for chess.
It can be speculated that in the general form the problem may play an important role in computer science. There are not to many optimization methods for EXPTIME complete
problems compared to the NP and P problems. It may be hoped that chess as a general problem may reveal some general methods for EXPTIME problems. Chess as a problem
may provide answers for fundamental questions about the limits of search optimizations for EXPTIME problems. The paper addresses to scientists and engineers interested in the topic.

	\subsection{The research methodology and  scenario}

	The research methodology is based on generalizing the method of partial depth and in the quantification of information gain in the exploration of the search space. The mathematical description of information in computer chess and its role in exploration is the central 
	idea of the approach. The method can be used to describe search also in other strategy games as well as in general. The problem is to quantify 
	the information gain in the particular state space where the search takes place.

Because the model used for describing search is interdisciplinary involving knowledge from several fields,
	a presentation of these areas is undertaken. Some knowledge from chess, game theory, information theory, computer chess 			algorithms, and previous research in the method of partial depth scheme are presented. Some of the important concepts
	in computer chess are modeled using information theory, and then the consequences are described. An implementation of the formula derived by the principles described in this theory of search based on information theory is presented along with results.  
	
	\subsection{Background knowledge}

	\subsubsection{The games theory model of chess}

	An important mathematical branch for modeling chess is games theory, the study of strategic interactions.
	
	\begin{definition}
		Assuming the game is described by a tree, a finite game is a game with a finite number of nodes in its game tree.
	\end{definition}
	
	It has been proven that chess is a finite game. The rule of draw at three repetitions and the 50 moves rule ensures that chess is a finite game. 

	\begin{definition}
		Sequential games are games where players have some knowledge about earlier actions. 
	\end{definition}

	\begin{definition}
		A game is of perfect information if all players know the moves previously made by all players. 
	\end{definition}

	Zermelo proved that in chess either player $\romannumeral 1$ has a winning pure strategy, player $\romannumeral 2$ has a winning pure strategy, or either player can force a draw.

	\begin{definition}
		A zero sum game is a game where what one player looses the other wins.
	\end{definition}

	Chess is a two-player, zero-sum, perfect information game, a classical model of many strategic interactions. 

	By convention, W is the white player in chess because it moves first while B is the black player because it moves second.
	Let M(x) be the set of moves possible after the path x in the game has been undertaken.
	W choses his first move $w_1$ in the set M of moves available.  B chooses his move $b_1$ in the set M($w_1$): $b_1$  $\in$ M($w_1$)
	Then W chooses his second move $w_2$, in the set M($w_1$,$b_1$): $w_2$  $\in$  M($w_1$,$b_1$) 
	Then B chooses his his second move $b_2$ in the set M($ w_1$,$b_1$,$w_2$): $b_2$ $\in$ M($w_1$,$b_1$,$w_2$) At the 			end, W chooses his last move $w_n$ in the set M($w_1$, $b_1$, ... ,$w_{n-1}$ ,$b_{n-1}$  ). \newline
	In consequence $w_n$ $\in$ M($w_1$, $b_1$, ... ,$w_{n-1}$ ,$b_{n-1}$  )

	Let n be a finite integer and M, M($w_1$), M($w_1$,$b_1$),...,\linebreak M($w_1$, $b_1$, ... ,$w_{n-1}$ ,$b_{n-1}$,$w_n$) 
	be any successively defined sets for the moves $w_1$,$b_1$,...,$w_n$,$b_n$ satisfying the relations: \newline
	
	\begin{equation}  b_n  \in  M(w_1, b_1, ... ,w_{n-1} ,b_{n-1},w_n  )  \label{EQ} \end{equation}    and \newline
	\begin{equation}  w_n  \in  M(w_1, b_1, ... ,w_{n-1} ,b_{n-1}  )         \label{EQ} \end{equation}    \newline   

	\begin{definition}
		A realization of the game is any 2n-tuple ($w_1$, $b_1$, ... ,$w_{n-1}$ ,$b_{n-1}$,$w_n$,$b_n$ ) satisfying the relations (1) and (2) 	
	\end{definition}

	A realization is called variation in the game of chess.

	Let R be the set of realizations (variations) , of the chess game. Consider a partition of R in three sets $R_w$ ,$R_b$ and $R_{wb}$ so that for any realization in $R_w$, player1 ( white in chess )  wins the game, for any realization in $R_b$ , player2 (black in chess) wins the game and for any realization in $R_{wb}$, there is no winner (it is a draw in chess).

	Then R can be partitioned in 3 subsets so that
	
	\begin{equation} R = R_w + R_b + R_{wb}   \label{EQ} \end{equation}    \newline  

	W has a winning strategy if $\exists$   $w_1$  $\in$   M ,  $\forall$  $b_1$  $\in$  $M(w_1)$  ,\linebreak $\exists$  $w_2$ $\in$ M($w_1$,$b_1$)  ,   $\forall$  $b_2$ $\in$ M($ w1$, $b1$, $w2$ ) ...\newline $\exists$ $w_n$ $\in$ M($w_1$,$b_1$,...,$w_{n-1}$,$b_{n-1}$), \linebreak $\forall$ $b_n$ $\in$ M($w_1$,$b_1$,...,$w_{n-1}$,$b_{n-1}$,$w_n$) ,
  where the variation
\begin{equation} ( w_1 , b_1 , \ldots , w_n, b_n ) \in R_w  \label{EQ}\end{equation} 

   	W has a non-loosing strategy if $\exists$   $w_1$  $\in$   M ,  $\forall$  $b_1$  $\in$  $M(w_1)$  ,\linebreak $\exists$  $w_2$ $\in$ M($w_1$,$b_1$)  ,   $\forall$  $b_2$ $\in$ M($ w_1$, $b_1$, $w_2$ )...\newline  $\exists$ $w_n$ $\in$ M($b_1$,$w_1$,...,$w_{n-1}$,$b_{n-1}$), \linebreak $\forall$ $b_n$ $\in$ M($w_1$,$b_1$,...,$w_{n-1}$,$b_{n-1}$,$w_n$) , where the variation
\begin{equation} ( w_1 , b_1 , \ldots , w_n, b_n ) \in R_w + R_{wb} \label{EQ}\end{equation}

	B has a winning strategy if $\exists$   $b_1$  $\in$   M ,  $\forall$  $w_1$  $\in$  $M(w_1)$  ,\linebreak $\exists$  $b_2$ $\in$ M($w_1$,$b_1$,$w_2$ )  ,   $\forall$  $w_2$ $\in$ M($ w_1$, $b_1$) ...\newline  $\exists$ $b_n$ $\in$ M($w_1$,$b_1$,...,$w_{n-1}$,$b_{n-1}$,$w_n$), \linebreak $\forall$ $w_n$ $\in$ M($w_1$,$b_1$,...,$w_{n-1}$,$b_{n-1}$) ,
  where the variation
\begin{equation} ( w_1 , b_1 , \ldots , w_n, b_n ) \in R_b  \label{EQ}\end{equation} 

   	B has a non-loosing strategy if $\exists$   $b_1$  $\in$   M ,  $\forall$  $w_1$  $\in$  $M(w_1)$  ,\linebreak $\exists$  $w_2$ $\in$ M($w_1$,$b_1$)  ,   $\forall$  $w_2$ $\in$ M($ w_1$, $b_1$) ...\newline  $\exists$ $b_n$ $\in$ M($w_1$,$b_1$,...,$w_{n-1}$,$b_{n-1}$,$w_n$), \linebreak $\forall$ $w_n$ $\in$ M($w_1$,$b_1$,...,$w_{n-1}$,$b_{n-1}$) , where the variation
\begin{equation} ( w_1 , b_1 , \ldots , w_n, b_n ) \in R_b + R_{wb} \label{EQ}\end{equation}

\begin{theorem}  
	Considering a game obeying the conditions stated above, then each of the next three statements are true:\newline
	(\romannumeral 1). W has a winning strategy or B has a non-losing strategy. \\
	(\romannumeral 2). B has a winning strategy or W has a non-losing strategy. \\
	(\romannumeral 3). If $R_{wb}$ = $\emptyset$, then W has a winning strategy or B has a winning strategy.   \\
\end{theorem}
 
	If $R_{wb}$ is $\emptyset$, one of the players will win and if  $R_{wb}$ is identical with R the outcome of the game will result in a draw at perfect play from both sides. It is not know yet the outcome of the game of chess at perfect play.\\

	The previous theorem proves the existence of winning and non-losing strategies, but gives no method to find these strategies. A method would be to transform the game model into a computational problem and solve it by computational means.  Because the state space of the problem is very big, the players will not have in general, full control over the game and often will not know precisely the outcome of the strategies chosen. The amount of information 
gained in the search over the state space will be the information used to take the decision. The quality of the decision must be a function
of the information gained as it is the case in economics and as it is expected from intuition.

	\subsubsection{Brief description of some chess concepts}

	\paragraph{The reason for presenting some concepts of chess theory. }  
Some of the concepts of chess are useful in understanding the ideas of the paper. Regardless of the level of knowledge and skill in mathematics without a minimal understanding of important concepts in chess it may be difficult to follow the arguments.
  It is not essential in what follows vast knowledge of chess or a very high level of chess calculation skills. However, some understanding of the decision process in human chess, how masters decide for a move is important for understanding the theory of chess and computer chess presented here. The theory presented here describes also the chess knowledge in a new perspective assuming that decision in human chess is also based on information gained during positional analysis. An account of the method used by chess grandmasters when deciding for a move is given in a very well regarded chess book. ~\cite{Kotov}.

\paragraph{Combination } A combination is in chess a tree of variations, containing only or mostly tactical and forceful moves, at least a sacrifice and  		resulting in a material or positional advantage or even in check mate while the adversary cannot prevent its outcome.    The following is the starting position of a combination.

	\includegraphics[width = 2in , height = 2in]{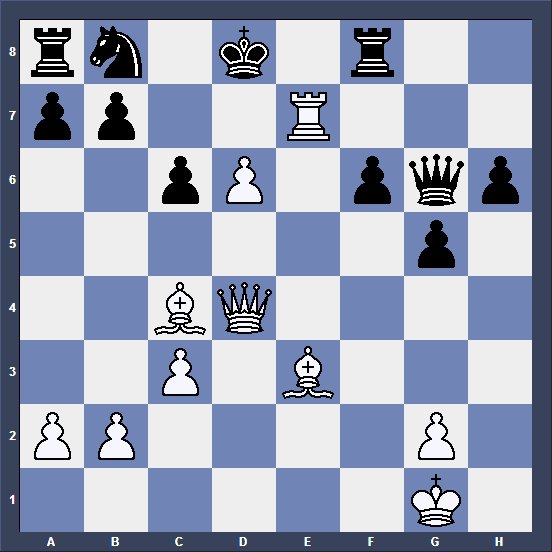}

The problem is to find the solution, the moves leading to the objective of the game, the mate.

	\paragraph{The objective of the game.} 
The objective of the game is to achieve a position where the adversary does not have any legal move and his king is under attack. For example a
mate position resulting from the previous positions is:

	\includegraphics[width = 2in , height = 2in]{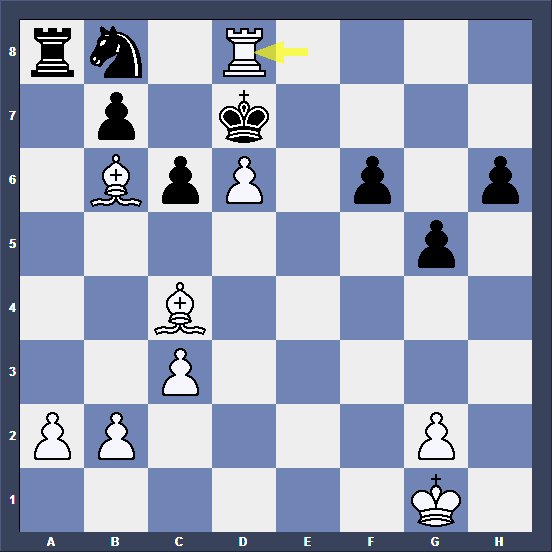}

	\paragraph{The concept of variation } 
A variation in chess is a string of consecutive moves from the current position. The problem is to find the variation from the start position to mate.

In order to make impossible for the adversary to escape the fate, the mate, it is desirable to find a variation that prevents him from doing so, restricting
as much as possible his range of options with the threat of decisive moves. 

	\paragraph{Forceful variation } 
A forceful variation is a variation where each move of one player gives a limited number of legal option or feasible options to the adversary, forcing the adversary to react to an immediate threat.

The solution to the problem, which represents also one of the test cases is the following:

1. Q - N6 ch! ; PxQ 2. BxQNPch ;  K - B1 3. R - QB7ch ; K - Q1 4. R - B7 ch  ;  k - B1  5. RxRch ;  Q - K1     6. RxQch ;  K-Q2      7. R-Q8  mate 

	\paragraph{Attack on a piece } 
In chess, an attack on a piece is a move that threatens to capture the attacked piece at the very next move.  
For example after the first move, a surprising move the most valuable piece of white is under attack by the blacks pawn.

	\paragraph{The concept of sacrifice in chess } A sacrifice in chess represents a capture or a move with a piece, considering that
	the player who performs the chess sacrifice knows that the piece could be captured at the next turn. If the player loses a piece without realizing the piece could be lost then it is a blunder, not a sacrifice. The sacrifice of a piece in chess
	considers the player is aware the piece may be captured but has a plan that assumes after its realization it would place the initiator in advantage or may even win the game. For example the reply of the black in the forceful variation shown is to capture the queen. While this is not the only option
possible, all other options lead to defeat faster for the defending side.  The solution requires 7 double moves or 14 plies of search in depth.

	\subsection{The axiomatic model of information theory}

	\subsubsection{Axioms of information theory}

	The entropy as an information theoretic concept may be defined in a precise axiomatic way. ~\cite{CoverThomas}.
	 
	Let a sequence of symmetric functions \( H_m( p_1 , p_2 , p_3, \ldots , p_m ) \) satisfying the following properties:\\
	(\romannumeral 1)  Normalization: \begin{equation}   H_2( \frac{1}{2} , \frac{1}{2} ) = 1    \label{EQ}\end{equation}
	(\romannumeral 2)  Continuity:\begin{equation} H_2(p,1 - p) \label{EQ}\end{equation}   is a continuous function of p \\        
	(\romannumeral 3) \begin{equation} H_m(p_1,p_2,...,p_m ) = H_{m-1}(p_1 + p_2 , p_3,...,p_m) = (p_1+ p_2)H_2(\frac{p_1}{p_1 + P_2},
\frac {p_2}{p_1 + p_2} )	\label{EQ}\end{equation}

	It results $H_m$ must be of the form 

	\begin{equation} H_m = - \sum_{x \in S} {p(x)*\log{ p(x)}}   \label{EQ}\end{equation}

	\subsubsection{Concepts in information theory}
	
	Of critical importance in the model described is the information theory. It is proper to make a short outline of information theory 			concepts used in the information theoretic model of strategy games and in particular chess and computer chess. 
	
	\begin{definition}
		A discrete random variable $\chi$ is completely defined by the finite set of values it can take S, and the probability 			distribution ${P_x(x)}_{x \in S}$. The value  $P_x(x)$ is the probability that the random variable $\chi$ takes the value x.
	\end{definition}
	
	\begin{definition}
The probability distribution $P_x$ :S $\rightarrow$ [0,1] is a non-negative function that satisfies the normalization 			condition \begin{equation}  \sum_{x \in S} P_x(x) = 1   \label{EQ}\end{equation}
	\end{definition}

	\begin{definition}
		The expected value of f(x) may be defined as 
		\begin{equation}  \sum_{x \in S} P_x(x)*f(x)   \label{EQ}\end{equation}
	\end{definition}

	This definition of entropy may be seen as a consequence of the axioms of information theory. It may also be defined independently ~\cite{CoverThomas}. As a place in science and in engineering, entropy has a very important role. Entropy is a fundamental concept of 
the mathematical theory of communication, of the foundations of thermodynamics, of quantum physics and quantum computing. 

	\begin{definition}
		The entropy $H_x$ of a discrete random variable $\chi$ with probability distribution p(x) may be defined as
		\begin{equation} H_x = - \sum_{x \in S} {p(x)*\log{ p(x)}}   \label{EQ}\end{equation}
	\end{definition}

	Entropy is a relatively new concept, yet it is already used as the foundation for many scientific fields. This article creates the 
	foundation for the use of information in computer chess and in computer strategy games in general. However the concept
	of entropy must be fundamental to any search process where decisions are taken.

	Some of the properties of entropy used to measure the information content in many systems are the following:

	\paragraph{ Non-negativity of entropy }

	\begin{proposition} 
		\begin{equation}   H_x \geq 0 \label{Prop}\end{equation}
	\end{proposition}

	\begin{interpretation} 
		Uncertainty is always equal or greater than 0.If the entropy, H is 0, the uncertainty is 0 and the random variable x takes a certain value with probability $P(x)$ = 1   
	\end{interpretation}

	\begin{proposition} 
		Consider all probability distributions on a set S with m elements.
	           H is maximum if all events x have the same probability, $p(x)$ = $\frac{1}{m}$ 	
	\end{proposition}

	\begin{proposition} 
		If X and Y are two independent random variables , then 
		\begin{equation}     P_{X,Y}(x,y) = P_x(x)*P_y(y)           \label{Prop}\end{equation}
	\end{proposition}

	\begin{proposition} 
		The entropy of a pair of variable X and Y is \begin{equation}  H_{x,y} = H_x + H_y  \label{Prop}\end{equation}
	\end{proposition}

	\begin{proposition} 
		For a pair of random variables one has in general 
		\begin{equation} H{x,y} \leq  H_x + H_y \label{Prop}\end{equation}
	\end{proposition}

	\begin{proposition} Additivity of composite events

		The average information associated with the choice of an event x is additive, being the sum of the information
		associated to the choice of subset and the information associated with the choice of the event inside the subset, 							weighted by the probability of the subset		

	\end{proposition}

	\begin{definition}
		
		The entropy rate of a sequence   $x_N$ =   $X_t$ , t $\in$ N
		\begin{equation} h_x = \lim_{N\to\infty}\frac{H_{x_N}}{N}   \label{def}\end{equation}

	\end{definition}

	\begin{definition}

	Mutual information is a way to measure the correlation of two variables

\begin{equation} I_{X,Y} = - \sum_{x \in S, y \in T} {p(x,y)*\log{ \frac{ p(x,y) }{  p(x)*p(y) }  }  }	\label{def}\end{equation}

	\end{definition}

	All the equations and definitions presented have a very important role in the model proposed as will be seen later in the article.

	\begin{proposition}

	\begin{equation}	I_{x,y} \geq 0	\label{def}\end{equation}

	\end{proposition}

	\begin{proposition}

		\begin{equation}	I_{X,Y} = 0 \label{def}\end{equation}   if any only if X and Y are independent variables.

	\end{proposition}

	\subsection{Previous research in the field}
	
	(\romannumeral 1) The structure of a chess program presented by Claude Shannon in ~\cite{Shannon} described the first model of a chess program. The following results of  ~\cite{Shannon} are fundamental. 

	\subparagraph{For a 1 move deep search: }  Let $M_i$ be the moves that can be made in position P and $M_{i}P$ denote the 
	resulting position when $M_i$ are applied to P. The solution is to chose $M_i$ that maximizes $f(M_{i} P)$
	\subparagraph{For a 4 move deep search} let $M_{ij}$ be the answer of black to the move of white, denoted as $M_i$ and so on.The formula is 
\begin{equation} max_{M_i} min_{M_{ij}} max_{M_{ijk}} min_{M_{ijkl}} f( M_{ijkl} M_{ijk} M_{ij} M_{i} P)  \label{EQ}\end{equation}

	(\romannumeral 2) The search extensions represent the interpretation  given by Claude Shannon to the way human chess masters solve the problem of following the forceful variations. 

	(\romannumeral 3) The quiescent search represents the solution to the problem of evaluating the positions with a static evaluation function given in ~\cite{Shannon} by Shannon.The idea is that after a number of levels of search a function
would perform only moves such as checks, captures, attacks.

	(\romannumeral 4) Following lines of high probability when analyzing positions represents the solution given by Claude Shannon to the selection of variations. ~\cite{Shannon}

	(\romannumeral 5) The result of Donald Knuth in regard to the connexion between the complexity of the alpha-beta 		algorithm and the ordering of the moves shows that when moves are perfectly ordered, the complexity of the search is the best possible for the method alpha-beta, corresponding to the best case possible. ~\cite{Knuth}

\begin{equation}  T(n) = b^{  \lfloor{  \frac{n}{2}} \rfloor  }   + b^{  \lceil \frac{n}{2} \rceil  } - 1  \label{EQ}\end{equation}	
	The complexity of alpha-beta for the worst case:  
	\begin{equation}  T(n) = b^{n}		\label{EQ}\end{equation}

	(\romannumeral 6) The idea of former world champion M.M. Botvinnik has been   to use the trajectories of pieces for the purpose of developing an intelligent chess program ~\cite{Botvinnik1} ~\cite{Botvinnik2} . The ideas of Botvinnink are important
because he has been a leading chess player and expert in chess theory. 

	(\romannumeral 7) A necessary condition for a truly selective search given by Hans Berliner is the following : The search follows the areas with highest information in the tree ~\cite{HBerliner2}  ``It must be able to focus
	the search on the place where the greatest information can be gained toward terminating the search''.
	Berliner describes the essential role played by information in chess, however he does not formalize the concept of information
	in chess as an information theoretic concept. From the perspective of the depth in understanding the decision process in chess
	the article  ~\cite{HBerliner2} is exceptional but it does not formulate his insight in a mathematical frame. It contains great chess and computer chess analysis but it does not define the method in mathematical definitions, concepts and equations.

	(\romannumeral 8) Yoshimasa Tsuruoka, Daisaku Yokoyama and Takasho Chikayama describe in ~\cite{Tsuruoka} a game-tree
	search algorithm based on realization probability.The probability that a move is played is given by the formula 

	\begin{equation}  P_c = \frac{n_p}{n_c}   \label{EQ}\end{equation}
	where $n_p$ is the number of positions in which one of the moves belonging to this category was played, and $n_c$ is the
	number of positions in which moves of this category belong.

Their examples are from Shoji but the method can be applied also to chess and deserves to be mentioned. They describe the realization probability of a node as the probability that the moves leading to it will actually be played. Their algorithm expands a node as long as the realization probability of a node is greater than a
threshold. They define the realization probability of the root as 1. The transition probability can be calculated recursively in the following way: 

	\begin{equation}             P_x = P_m * P_{x \prime}        \label{EQ}\end{equation}

	where $P_m$ is the transition probability by a move m, which changes the position ${x \prime}$ to x.   $P_x$ is the realization probability of node x, and $P_{x \prime}$ is the realization probability of parent node ${x \prime}$. 
The decision if to expand or not a node is given by this rule. The probability of a node gets smaller with the search depth in this method because transition probabilities are always smaller than 1. The node will become a leaf if the realization of a node is smaller than a threshold value. The method has been implemented by adding the logarithms of the probabilities. In this method, when there is just one move, the transition probability will be 1. The transition probabilities are also determined by the category the move belongs to. Categories are specific to the game of Shoji and are similar to chess to some extent: checks, capture, recapture, promotion and so on. When a move belongs to more than one category, then the highest probability is taken into account.
If there are multiple legal moves from a category, the probability that one move is chosen is smaller than the probability of the category. The probability of a move is taken from real games.

	(\romannumeral 9)  Mark Winands in ~\cite{MWinands} outlines a method based on fractional depth where the fractional ply FP of a move with a category c is given by 
	
		\begin{equation}    FP = \frac{\lg P_c}{ \lg C }   \label{EQ}\end{equation}

	His approach is experimental and based on data mining as the method presented previously.

	(\romannumeral 10) In the article ~\cite{SEX}  David Levy, David Broughton, Mark Taylor describe

	the selective extension algorithm. The method is based on ''assigning an appropriate additive measure for the interestingness of the terminal node'' of a path.

	Consider a path in a search tree consisting of the moves $M_1$, $M_{ij}$, $M_{ijk}$ and the resulting position being a 
	terminal node. The probability that a terminal node in that path is in the principal continuation is

	\begin{equation}   P( M_i )*P( M_{ij} )*P( M_{ijk} )     \label{EQ}\end{equation}

	The measure of the ''interestingness'' of a node in this method is

	 \begin{equation}      lg[  P( M_i ) ] + lg[ P( M_{ij}  ] + lg[ P( M_{ijk} )  ]     \label{EQ}\end{equation}

	\subsection{analysis of the problem}

	The problem is to construct a model describing the search process in a more fundamental way starting from axioms, possibly in an informational theoretic way and derive important 
	results known in the field. In this case shall be described the elements of the search based on informational theoretic 				concepts. The player who is able to gain most information from the exploration and calculation of variations will take the most informed decision, and has the greatest chance to win. It is very likely that
the skill of human players consist also in gaining most information from the state space for taking the best decision. In this case the human decision and its quality is expressed by its economical reality, the better informed decision-maker has the upper hand.

	\subsection{Contributions}	
 
	The contribution of the model presented here is aimed to establish a mathematical foundation for computer chess and in general for 
	computation of strategic decisions in games and other fields. The model describes the uncertainty of a position through
	the mathematical concept of entropy and derives important consequences. Some of these consequences have been established through different 
	other methods. Here are presented in the context of the information theoretical model of computer chess. A new algorithm, based on the idea
	of directing search towards the lines of highest information gain is presented. The algorithm proposed is based on the model described in the paper.
	In this way it is proven that using almost no specific chess knowledge a simple scheme gives significantly better results than a ordinary alpha-beta using comparable power.  Other results used empirically or on different other grounds
	before are presented as consequences of the model introduced here. The consequences are shown in the result section. 

 The article establishes a 				mathematical foundation for quantifying the search process in computer chess,  based on the axioms of information theory and the concept of entropy. The 	parameter that controls the depth of search is linked to the fundamental basis of the information theory. In this way some of 
	the most important concepts of computer chess,  are described by mathematical concepts and measures. This approach
	can be extended to describe other important results in computer chess in special and in games in general.
 
	If for the 8x8 particular case the intuitive approach has been sufficient, for describing in a scientific way the general NxN 			chess problem  it is more likely that a fundamental mathematical model will have much more explanatory power. 

	The concept of information gain used in other areas of artificial intelligence is used, maybe for the first time in computer chess
	to describe the quality of the moves and their impact on the decrease in the entropy of the position.
	The paper proposes a new model , representing a new way of looking at computer chess and at search in artificial intelligence in general. It shows the effectiveness and the power of the model in explaining a wide range of results existing in the field and also to show new results. The model is characterized by novelty in looking to the problems of chess in its scientific dimension. An in depth presentation of the model is given including extensive 
background information. Many of the most important known facts in computer chess are presented as consequences of model. A quantitative view on the architecture of the 
evaluation function is given, opening the way to proofs about the limits of decision power in chess and in other games.

	\section{Search and decision methods in computer chess}

	\subsection{The decision process in computer chess}

	The essence of the decision process in chess consist in the exploration of the state space of the game and in the selection between the competing
	alternatives, moves and variations. The amount of information obtained during exploration will be a decisive factor in 
	a more informed decision and thus in a better decision. It is the objective the exploration process to find a variation as close as possible to the optimal minimax variation. The player finding a better approximation for the minimax perfect line will likely deviate less from 
the optimal strategy, will control the game and therefore gain advantage over the other player.

	\subsection{Search methods}

	\subsubsection{algorithmic and heuristic search in strategy games}

		In its core objective the minimax heuristic searches for approximate solutions for a two player game where no player has anything
	to gain by changing his strategy unilaterally and deviating from the equilibrium. The objective of the application of information theory to chess would
	be to orient the search on the lines with highest information gain. This could result in the minimax method taking a more informed 
	approach. The search
	process has as objective to gain more information about the exact value and to reduce the uncertainty in the evaluation of the 		position for the player undertaking the search. Therefore the player or decision-maker who uses a search method capable of 			gaining more 			information will take the decision having more information and will have a higher chance to win. The player who has better 			information due to a heuristic capable of extracting more information from the state space will very likely deviate less from the 			minimax strategy and will likely prevail over a less informed decision-maker.

	\subsubsection{the alpha-beta minimax algorithm} 
	 The paper of Donald Knuth  ~\cite{Knuth} contains an illustrative implementation of minimax. This may be considered also an implementation of Shannon's idea. The procedure minimax can be characterized by the function
$$ 
F(p) = \left\{ \begin{array}{rl}
	F(p) = f(p) &\mbox{ if $d = 0$  } \\
	  max( -F(p_1) ,...., -F(p_d)    )	 & \mbox{ if $d > 0$ }
		\end{array} \right.
$$

	These classic procedures are cited for comparison with the methods where the depth added with every ply in search is not always 
1 but may be less than one in the case of good moves, or more than one in the case of less significant moves. Consider for example the procedure F2 as described in the classic paper of Donald Knuth, implementing alpha-beta,  ~\cite{Knuth} and the G2 procedure which assumes a bounded rationality.

Knuth in ~\cite{Knuth} proves that the following theorem gives the performance of alpha beta for the best case:

\begin{theorem}

	Consider a game tree for which the values of the root position is not  $\pm$ $\infty$ , and for which the first successor of every position is optimum.

	If every position on levels 0,1,..,,n-1 ,  has exactly d successors, for some fixed constant d, then the 
	alpha-beta procedure examines exactly

\begin{equation}  T(n) = b^{  \lfloor{  \frac{n}{2}} \rfloor  }   + b^{  \lceil \frac{n}{2} \rceil  } - 1  \label{EQ}\end{equation}

	positions on level n

\end{theorem}

	\paragraph{Search on informed game trees \newline} 

	In ~\cite{PijlsBruin} it is introduced the use of heuristic information in the sense of upper and lower bound but no reference to any information theoretic concept is given. Actually the information theoretic model would consider a distribution not only an interval as in ~\cite{PijlsBruin}.
	Wim Pijls and Arie de Bruin presented a interpretation of heuristic information based on lower and upper estimates for a node and integrated it in alpha beta, proving in the same time the correctness of the method under the following specifications.  

Consider the specifications of the procedure alpha-beta.
If the input parameters are the following: \newline
(1) n, a node in the game tree,               \newline
(2) alpha and beta , two real numbers and  \newline
(3) f , a real number, the output parameter, \newline
and the conditions: \newline
(1)pre:   alpha $<$ beta  \newline
(2)post: \newline	
	alpha $<$ f $<$ beta   $\Longrightarrow f = f(n)$,                         \newline
	f $\leq$  alpha	$\Longrightarrow$ f(n) $\leq$  f $\leq$  alpha  \newline
	f $\geq$  beta  $\Longrightarrow$ f(n) $\geq$ f $\geq$ beta          \newline

then

\begin{theorem}

	The procedure alpha-beta (defined with heuristic information, but not quantified as in information theory)   meets the specification.~\cite{PijlsBruin}

\end{theorem}

	Considering the representation given by  ~\cite{PijlsBruin}, assume for some game trees, heuristic information on the minimax value f(n) is available
	for any node. 

          \begin{definition}
	              The information may be represented as a pair H = (U,L), where U and L map nodes of the tree into real numbers.
          \end{definition}

	\begin{definition}
	           U is a heuristic function representing the upper bound on the node. 
	\end{definition}

	\begin{definition}
	           L is a heuristic function representing the lower bound on the node.
	\end{definition}

	For every internal node, n the condition U(n) $\geq$ f(n) $\geq$ L(n) must be satisfied. \newline

	For any terminal node n the condition U(n) = f(n) = L(n) must be satisfied. This may even be considered as a condition for a leaf.

	\begin{definition}
		A heuristic pair H = (U,L) is consistent if  \\
	    	U(c) $\leq$ U(n)      for every child c of a given max node n and  \\
	  	L(c) $\geq$ L(n)      for every child c of a given min node n 
	\end{definition}

	The following theorem published and proven by ~\cite{PijlsBruin} relates the information of alpha-beta and the set of nodes visited.

	\begin{theorem}

		Let $H_1$ = ($U_1$,$L_1$) and $H_2$ = ($U_2$,$L_2$) denote heuristic pairs on a tree G, such that $U_1$(n) $\leq$ $U_2$(n) and $L_1$(n) $\geq$ $L_2$(n) for any node n. Let $S_1$ and $S_2$ denote the set of nodes, that are visited during execution of the alpha-beta procedure on G with $H_1$ and $H_2$ respectively, 
then $S_1$ $\subseteq$ $S_2$.

	\end{theorem}

	\section{The information theoretic model of decision computer chess}

	\subsection{The intuitive foundations of the model }

It is a well known fact, in computer chess various lines of search do not contribute equally to the information used for deciding moves. The model shows why certain patterns of exploration result in a more informed search and in a more informed decision. The use of stochastic modeling in computer chess 
does not imply the game has randomness introduced by the rules but by the limits in predicting and controlling the variables used for modeling the 
search process. The object of analysis is not chess or other game but specifically the random variables used in the systems and the search heuristics capable of taking decisions in chess and other strategy games. Many or even all modern algorithms in computer chess are probabilistic. A few examples are the B* probability based search  ~\cite{HBerliner1}  ~\cite{HBerliner2} and the fraction ply methods published in  ~\cite{Tsuruoka} ~\cite{SEX} ~\cite{MWinands}. This articles describe the decisions such as the continuation or braking of a variation or the selection of nodes as probabilistic. Even if some of the previous cited articles do not describe a stochastic process or system,  is is possible to define the methods as part of such systems or within the general principles of such systems. It is natural in this framework to describe the variations as stochastic processes.

	\subsection{System analysis:  random variables in computer chess}

	Chess as a deterministic game apparently does not have random variables. Yet the systems deciding chess moves without exception use random 
	variables. If for the 8x8 chess some day in the future there may be possible to construct systems that do not use any random variable, for 
	the general NxN problem assuming there will always be systems capable of infinite computational power is not feasible. Therefore a better solution
	would be to analyze the problem assuming the uncertainty is not removable because the size of the system is infinite. 

	Some of the critical variables of the system are the trajectories of pieces, the move semantics, the values of positions along a variation, the 			evaluation error. These variables could be defined in the following way:
	
	\begin{definition}
		The trajectory of a piece is the set of successive positions a piece can take. The uncertainty  in regard to the position of 			the 			piece during the 			search process, given a heuristic method can be seen as the entropy of the trajectory $H_{trajectory}(p)$. 
	\end{definition}

	If the heuristic method is simple it may be guessed something about the trajectory, but if the search implements 6000 - 10000 knowledge elements and 
	many heuristics the process for various lines will be marked by uncertainty on the scale of individual variables along a search line but may be 			controllable at the scale of the entire search process. If no assumption is made on the principles or knowledge of the game this can be 			described as a random walk. 

	\begin{definition}

		The move semantics can be defined as the types of moves and the way they combine in chess combinations and plans. It may be defined an 				uncertainty in regard to the  semantics of strategic and tactical operations in chess in terms of the chess categories of moves $H_c(p)$ . 

	\end{definition}

	\begin{interpretation}

		The strings of moves, captures, checks, threats are like an alphabet of symbols. These symbols are the alphabet of the chess strategy and 
		tactics. The patterns present in combinations are the ideas constructed with these symbols. In this way it is shown here the mathematical model 				and theory that supports the reality expressed by masters, that behind each combinations is an idea. 

		The entropy o the alphabet of chess tactics and strategy can be described in terms of the entropy of a string of moves with their respective classes in the same way it is described the entropy of an alphabet and its symbols. A description will be shown in the context of chess.

	\end{interpretation}

	The error resulted from the application of the evaluation function on a position can be described as a random variable having associated an 
	uncertainty $H_e$ , uncertainty in regard to the error.
	
	The fluctuations of positional value resulted by the alternative minimizing and maximizing moves may be described as a random walk if the 
	game is balanced. In any case an uncertainty $H_s$ may be defined in regard to the result of the search process, as long as it cannot be predicted the result in a certain way.  

	\subsection{The mathematical foundations of the model}

	\subsubsection{The quantification of the uncertainty in the value of a position}

		\paragraph{preliminary analysis:}

	The value of a chess or other game position may be represented in different ways.

	(\romannumeral 1) Representation using the +1/0/-1 values as described by game theory.
 
	The value of the position may be considered a measure of the quality of a certain position. 

	\begin{equation} f(P) = 	 +1 \label{EQ}\end{equation}	for a won position, 
           \begin{equation} f(P) =   	   0 \label{EQ}\end{equation}	for a drawn position, 
           \begin{equation} f(P) = 	  -1 \label{EQ}\end{equation}	for a lost position.

	(\romannumeral 2) Representation of the value of a position using a real or integer number and an interval

	However a more general method is to assign an integer or real value as a measure of the probability of a node having the above 	mentioned values.
The range of the evaluation function may be described by an interval. The closer to the limits of the interval a value is the more likely in this model a position is to have a value
close to the perfect game theoretical values  +1/0/-1  .  The above mentioned values +1,-1,0  could be recovered as a particular case of a real value approach.This representation is probably the most used in computer chess and other games.

	(\romannumeral 3) Representation of the value of a position using a distribution

 The representation of positions value in chess as seen by world champion Hans Berliner: ``The value of an idea is represented by a probability distribution of the likelihoods of the values it could aspire to.This representation guides a selective search and is in turn modified by it.'' ~\cite{HBerliner2}
Therefore Berliner expresses the idea of a system in a qualitative way. However he does not elaborate on a quantitative description and consequences.
The articles  ~\cite{HBerliner1} ~\cite{HBerliner2} describe the B* algorithm but in a qualitative way and do not make use of a possible mathematical 
description for this idea. 
It is possible a mathematical quantification of the idea described by the former world champion.

	(\romannumeral 4) Representation of the value of a position using the information theoretic concept of entropy \newline
The contribution of this article is in ,a mathematical model describing the decision in computer chess, in chess and the knowledge of chess in a integrated theory. This mathematical representation proposed here could generalize
the chess tactics and strategy, developed by chess masters for the 8x8 case to a NxN general model, describing strategy and tactics in a mathematical 
theory and finding the 8x8 case as a particular case.

	In this way, the methodsl described previously can be generalized by representing the value of a position as a distribution with the uncertainty associated modeled as the entropy of the search process.

	\begin{equation} H_{value}(position) 	=  \sum_{i=1}^{\infty}{P_i}log P_i  \label{EQ}\end{equation}	

	where $P_i$ is the probability of the position taking a certain value.

	(\romannumeral 5) Representation of the quality of a variation using a semantical model

	It may be defined based on the type of operations, moves such as check, capture, attack and so on. Many of the moves do not have a classification with a particular
	name , but significant moves usually have. It may be possible that the range of possible semantics for moves is far greater than the known categories.
	It is reasonable to consider that the range conceivable could be even greater for the NxN chess. Limiting the analysis to the classical 8x8 game of chess it may be observed
	from the previous chess problem, the combination,that significant variations are often composed of significant moves such as those from the above
	mentioned categories. In combinatorial positions the variations leading to victory are overwhelmingly composed of forceful moves such as checks 			and captures. Any book with combinations , for example 1001 Brilliant way to checkmate will reveal that combinations have almost only such moves
	and often start with a surprising move such as the sacrifice of a chess piece. The number of lines with checks differs in practice from position to 
	position, however from the total number of moves in combinatorial positions usually less than 20 \% are checks and captures but probably these 20 \% account
	for something like 80\% of the victories in decisive complex positions. In the positions selected from books on combinations the percentage is not 80 \% but 100\% , practically each and every position in 1001 Brilliant way to checkmate by Fred Reinfeld is so . From the entire number of variations  , forceful lines with checks and captures account for maybe 1\%
	 but something like 99\% of victories in decisive combinatorial positions. In this way it can be justified the old say in chess that
	''Chess is 99\% tactics''. Therefore such considerations of semantic nature decrease the uncertainty on the decision to investigate a variation very much.
	The categories of moves and the semantic of variations explain why for such line the uncertainty is much smaller than
	for ordinary lines of play. For such lines the probability to be on the principal variation is much higher that for other lines.

	 For such a line the uncertainty in regard to the possibility that such string of moves is a principal variation is much lower than for normal lines.

	\begin{equation} H_{semantic}(PATH) 	=  \sum_{i=1}^{\infty}{P_i}log P_i  \label{EQ}\end{equation}	

	where $P_i$ is the probability of a PATH to a position containing a sequence of moves with such categories to be on the principal variation. As one can see from chess analysis it is also much more likely that good players analyze such lines than ordinary variations with no semantic meaning. An idea is 
composed of a sequence of symbols . The ideas in chess must be composed from an alphabet of symbols. The mathematical model constructed describes its properties by defining the entropy associated with it and its meaning in terms of chess.

This may be considered the mathematical description of the 
	expression ''lines of high probability'' as Shannon calls in an intuitive way such variations, but does not offer any mathematical construct to model this. There was
certainly no experimental basis for it at that time.
	The right model that he did not use to describe the lines of high probability in chess mathematically may be actually his theory of information. 
           Probably in his time, computer chess was a field to new and there were not yet the facts needed to make this generalization.

This article aims    to advance the level of knowledge and make this generalization now in the form of a model of decision in chess based on information theory. 

Many of the facts known in chess and computer chess can be explained through the information theoretic model therefore the data already known providing an excellent verification of this new description.

	\subsubsection{The quantification of the uncertainty of a trajectory in state space}

	The search process may be represented in computer chess and other areas as a process of reducing the uncertainty in the mathematical sense, assuming the trajectories of pieces are modeled as random walks or random walks biased towards the most likely lines used in practice. This small change of perspective could produce large changes in the field of 
computer chess. Like in many areas of science, small changes may result in big effects.  	
	
	The uncertainty about a chess position is our lack of knowledge and prediction power about the change in some fundamental variables used to model search in chess and computer chess including the positions of pieces and the knowledge  about the perfect value of a position.The objective of search in computer chess could be described as 
the decrease in the uncertainty on important variables used to model a position. This includes also their value.

	The idea is to describe in a model, based on information theory essential elements of chess and computer chess such 		as the uncertainty in evaluation of positions, 		the effect of various moves in the variation of the entropy of a 		position, the entropy of various pieces, the information gain in the search for a move performed by a human player and
in computer chess search. The connexion 		between combinations, tactics and 			information theoretic concepts is clear in this model.
Human decision-making in chess can be described by the laws of economics but there is not much work in the area. Here a clarification is given. Because information is essential also in human decision-making as described by economics , the information gain over investigation of variations is what determines the quality of human decision making in chess.
The positional patterns perceived by humans can also be seen from their attribute of decreasing uncertainty in the positions value or predictable trajectory or the expected error in evaluation.

\begin{definition}

	A trajectory of a chess piece is a set of consecutive moves of a piece on the chess board. Also the entire board with all the pieces has a 
	trajectory in a state space. This is called variation in chess. 

\end{definition}

\begin{definition}

	A variation may be defined as a concatenation of the trajectories of pieces.

\end{definition}

	The search process can be described by a stochastic process where variables determined by the heuristic search method, such as the trajectory of pieces and the values returned by the evaluation function are unpredictable or have a significant degree of randomness . A variation in chess can be described or modeled as a stochastic process
in the context of a heuristic generating that variation. A trajectory of a piece in a variation may also be described by a stochastic process.

	Let p be a position and  $H_{trajectory}(position)$ be the entropy associated with the variations generated by the heuristic in the position. 

	\begin{equation} H_{trajectory}(position) 	=  \sum_{i=1}^{\infty}{P_i}log P_i  \label{EQ}\end{equation}	

	where $P_i$ is the probability of a certain trajectory in the state space.

	In the context of computer chess it is clear in the case of positions where it is not know the perfect value and it must be relied on an estimate, such representation must express in the best way possible the uncertainties about the possible outcomes of the position. Not only a variation or a trajectory may be described by random variables, but also the values of the positions in a variation. Even 
if it had been available the computation power capable of exploring all the consequences of the position, its value could still be expressed as a distribution if it is
considered the quality of a variation not only based on its value but also based on the length of the path to that final value +1/0/-1. This has also a 
practical meaning, because a lost position in absolute terms may not be lost if the path to defeat is long, complicated and the adversary may not find that path. There are plenty of end-games where the perfect value is known but many humans have a hard time achieving the perfect value.

	This could be a general description of the uncertainty of a position, not only in chess and computer chess but also in other strategy games and also
in heuristic search in general.

There is a second method to describe the uncertainty in the position. 

In order to determine how entropy changes after moves such as captures, which are known from practice to lead to less uncertainty, it can be seen that the number of possible combinations with the pieces left after the exchange is smaller so it results in a decrease in the entropy of the position.
It may be analyzed if this decrease can be quantified, in order to determine the direction where the search becomes more accurate. One method would be to calculate how many configurations are possible with the pieces existent on the board before and after the capture or exchange.

A position in chess is composed from the set of pieces and their placement on the board. The number of combinations possible with these pieces is often
very big, however, the number of positions that can be attained is much smaller.Many of these configurations would be impossible according to the rules of chess, other configurations would be very unlikely, and certainly the number of configurations significant for the actual play and close to the best minimax lines of the two players is even smaller. The number of positions that usually appear in games is even smaller but 
still significant. Therefore we have to look for a different metric for the decrease in entropy during combinations and other variations with many exchanges.

Instead of considering the combinatorial effects at the level of the number of positions or random moves, it could make more sense to represent the combinatorial element of the game at the level of trajectories. 
The number of moves possible along such trajectories is much smaller and in consequence the number of possible trajectories of pieces even smaller.

As a method of programming chess this has been already proposed by the former world 
champion, Botvinnik. He proposed it as a heuristic for a chess program but not in the context of information theory and in a context different from
the idea of this article. He used his intuition as a world champion, we try to formalize this mathematically.
It is rumored that many strong chess programs and computers, including Deep Blue, use trajectories from real games stored in databases as patterns for the move of pieces. This is already a strong practical justification for using trajectories of moves in a theoretical model.
The uncertainty along thr trajectories of pieces can be used to describe the information theoretic model of chess and computer chess.

Because each piece has its own trajectory, this idea justifies the assumption:

\begin{assumption}
The entropy of a position can be approximated by the sum of entropy rates of the pieces minus the entropy reduction due the strategical configurations. 
\end{assumption}

This can be expressed as:

	\begin{equation} H_{trajectory}(position) 	=  \sum_{i=1}^{N} H_{p_i} -  \sum_{i} H_{s_i}   \label{EQ}\end{equation} 

where $H_i$ represents the entropy of a piece and $H_{s_i}$ represents the entropy of a structure with possible strategic importance. 

This gives also a more general perspective on the meaning of a game piece. A game piece can be seen as a stochastic function having the state of the board as entrance and 
generating possible trajectories and the associated probabilities. These probabilities form a distribution having an uncertainty associated. 

The entropy of a positional pattern, strategic or tactic may be considered a form of joint entropy of the set of variables represented by pieces positions and their trajectory.
The pieces forming a strategic or tactic pattern have correlated trajectories which may be considered as forming a plan.

\begin{equation}H(s_i) =  - \sum_{x_i} ... \sum_{x_n} P(s_i) \log [P(s_i)]	 \label{EQ}\end{equation}

\begin{equation} H_{s_i} =  H(s_i) \label{EQ}\end{equation}

where $s_i$ is a subset of pieces involved in a strategic pattern and the probabilities $P_i$ represent the probability of realization of such strategic or tactical pattern. The reduction of entropy caused by strategic and tactical patterns such as double attacks,pins, is determined by both the frequency of such 
structures and by the significant increase in probability that one of the sides will win after this position is realized.

We may consider the pieces undertaking a common plan as a form of correlated subsystems with mutual information I(piece1,piece2,...). It results
that undertaking a plan may result in a decrease in entropy and a decrease in the need to calculate each variation. It is known from practice that planning decreases the need to calculate each variation and this gives an experimental indication for the practical importance of the concept of entropy as it is defined here in the context of chess . Each of the tactical procedures , pinning, forks, double attack, discovered attack and so on, can be understood formally in this way. A big reduction of the uncertainty in regard to the outcome of the game 
occurs, as the odds are often that such a structure will result in a decisive gain for a player.
When such a structure appears as a choice it is likely that a rational player will chose it with high probability. 

The entropy of these structures may be calculated with a data mining approach to determine how likely they appear in games.

An approximation if we do not consider the strategic structures would be: 

\begin{assumption}                                     
 
	\begin{equation} H_{trajectory}(position) 	=  \sum_{i=1}^{N} H_{p_i}  \label{EQ}\end{equation} 

\end{assumption}

\paragraph{assumption analysis:} The entropy of the position is smaller in general than the sum of the entropies of pieces because there are certain positional patterns such as openings, end-games, various pawn configurations  in a chess position which result in a smaller number of combinations, results in order and a smaller entropy. Closer to reality would be such a statement:

	\begin{equation} H_{trajectory}(position) 	\leq  \sum_{i=1}^{N} H_{p_i}  \label{EQ}\end{equation} 

Considering many real games we can estimate the decrease in the number of possible combinations and implicitly in entropy after a capture. This assumption is supported by the following arguments in favor of the model: \newline 

(\romannumeral 1)  It is much more similar to the way planning takes place in chess. Long range planning and computer chess is a good source for chess knowledge interpreted for computer chess. ~\cite{Botvinnik1} ,  ~\cite{Botvinnik2} \newline

(\romannumeral 2) The space of trajectories includes reachable positions as in a real game \newline  

(\romannumeral 3) The trajectories method gives a good perspective on the nature of intuition and of visual patters in chess. Before analyzing in a search tree, players see the possible trajectories on the board. \newline 

(\romannumeral 4) Taking in account trajectories of pieces results in the variations being a concatenation of trajectories and this is much more similar to what are most of the good variations in computer chess. A concatenation of trajectories is much more ordered and less entropy prone than the attempt to use all sorts of moves in a variation.   Actually the geometrical method of analogies concatenates search lines. ~\cite{Caissa} 
 
The geometrical method of Caissa does not rely on stored trajectories but it would be possible to concatenate trajectories already played.
While this method is very good for practical play on the 8x8 board, it may not be the most elegant as a method for theoretical analysis, because it is limited 
to the 8x8 case and to variations played until now. For the more general case, the NxN chess the calculation of the decrease in entropy after 
captures by using variations played on 8x8 chess is not possible. Therefore a more general methods must be used. 

A third method, the most general would consider the trajectories of pieces as a random walk on the chess board and the previous formula for calculating the entropy of a position. The trajectories of the pieces can be modeled mathematically in an approximative way using the model of random walks on the chess board. This method does not make any assumptions
on the style of play, openings, patterns of play used until now,or the fact that it is used a 8x8 board. The analysis can be used also for the NxN board. 

A first step would be to calculate the entropy of each piece.  The entropy of a chess piece can be calculated based on the idea of possible random walks from a position on the board. 

\begin{assumption}
	The random walk model approximates the model of trajectories of pieces om the chess board.
\end{assumption}

\paragraph{Analysis of the random walk assumption} The random walk model of describing the trajectories of pieces has several proved advantages over the case when all possible moves are taken into account:   \newline
(\romannumeral 1) It is more similar to the way humans visualize moves on the board before precise calculation.               \newline
(\romannumeral 2) It models very well the patterns of chess pieces consistent with averaging over many games.              \newline
(\romannumeral 3) The decision of chess players or programs is not random for a good player but the search process for a move in both human and 
machine has a lot of randomness. A computer has to analyze so many positions because many of the positions are meaningless, therefore the search 
has to some extent a random nature. However the evaluation and the decision are not random but based on more precise rules, usually minimax or approximations of minimax. \newline

	\subsection{The calculation of entropy rates of pieces}

The moves of a piece on the chess board can be described as a random walk, if we do not make assumptions about any particular knowledge extracted from chess games such as high probability trajectories of pieces in circumstances such as openings or tactical or strategical structures.The assumption of the random walk of pieces makes the model presented less rigid than the other options presented before and does not place any demand for top expert knowledge or any assumption related to data mining. Even if we consider the theory of chess, there are no precise rules on how to perform the search. The random walks model is more general and is feasible in the analysis of the NxN chess problem. The idea of modeling trajectories as random walks makes possible the extension of the information theoretic model of chess to programming GO. GO may also be programmed 
using random walks on the board using monte carlo methods. While on the 8x8 problem, expert opinions count, for the general problem, the NxN problem, there are no expert opinions.

A slightly modification of the idea of the random walk on a chess board is the idea of a random walk on a graph. A description of random walk of pieces on 
a graph, outside the context of this research but as an example of information theory is given in   ~\cite{CoverThomas}

The probabilities of transitions on such a graph are given by the probability transition matrix.

\begin{definition}
	A probability transition matrix  [$ P_{ij}$ ] is a matrix \newline defined by 
  $P_{ij} =  Pr \{ X_{n+1} = j  |  X_n = i  \}$
\end{definition}

The path of any piece on the chess board may be considered as a random walk on a graph or a biased random walk towards high probability trajectories. Consider now a model of the random walk of a chess piece on a chess board as a random walk on a weighted graph with m nodes. Consider a labeling of the nodes {1,2,3,...,m}, with weight $W_{i,j}$  $\geq$ 0 , on the edge joining node i to node j.  

\begin{definition} 
	An undirected graph is one in which edges have no orientation. If the graph is is assumed to be undirected, $w_{i,j}$ = $w_{j,i}$.
\end{definition}

\begin{assumption}
	The graph is assumed to be undirected, $w_{i,j}$ = $w_{j,i}$ .
\end{assumption}

	\paragraph{analysis of the assumption}: This assumption is largely verified in chess, and it is true in regard to the moves of all pieces, with the exception of pawn moves. However, any probabilistic assessment on the number of possible configurations would likely have to make the same assumption in regard to pawns movement so there is no
particular disadvantage of the method proposed in this regard. The castle is also a structure which decreases the entropy of the position.

	If there is no move between two positions, there is no edge joining nodes i and j, and $w_{i,j}$ is set to 0. 

	Consider a chess piece at vertex i. The next vertex j is chosen among the nodes connected to i with a probability proportional to the weight of the edge connecting i to j. In a probabilistic scenario \begin{equation}	 P_{ij} =  \frac{w_{i,j}}{\sum_{k} w_{i,k}}	\label{EQ}\end{equation}
	If we include knowledge on trajectories, from real games, then we could use the probabilities matrix and create a biased random walk. In this 
calculation it will not be used a biased random walk, but it is clearly possible to do so.It will be assigned a probability not empirically but resulted from the
number of connections with other nodes. The stationary distribution for this Markov chain should assign to node i a probability proportional to the total
weight of the edges emanating from i. The calculation of the entropy associated with pieces appears as an example for elements of information theory  
in  ~\cite{CoverThomas}  but not in the context of computer chess or related to any result from computer chess or as a proposal for any algorithm. Just as a very good example of information theory. 

\begin{definition}

The entropy rate or source information rate of a stochastic process is, informally, the time density of the average information in a stochastic process.

\end{definition}

\begin{analysis}

The interpretation in chess of this stochastic process is the trajectory, actual and possible of the pieces during the search process.
This is important, because we discuss here the trajectories during the search process not only what practically happens, the real trajectories in the game.
Indeed, nobody can say precisely where the decision to optimally break the variation will occur and what is the trajectory of the piece until that point, or 
what is the trajectory of the piece in the optimal line.

\end{analysis}

The entropy rate of various pieces is calculates in the above mentioned source. In the general form of the game, on a NxN board, the entropy rate is
for king = $\log$ 8 bits, for rocks is $\log$ 14 bits , for queen is $\log 28$ for bishop is $\log 14$ for knight is $\log 8$  .
\paragraph{analysis:} As it can be observed, the number of moves is a critical factor in the quantification of the uncertainty related to a chess piece.
A constant in front of the logarithm is necessary because of the edge effects. This constant is different for boards of different sizes. 
In chess, the general principles of the game sometimes do not explain some positional features. It may be conceived that edge effects are significant in
the 8x8 chess board problem. 

	\subsection{The entropy of the value of the position}

Let V be a random variable representing the values returned by the search process. As long as we cannot predict the values this is a random variable.
If we could predict the value, the search would be meaningless. 

So we can define an entropy related to the estimated value of each position of a variation in the search process. 
At the beginning of the game the value of the game is 0. During the game it deviates from this value. The deviation is measured by the evaluation 
function. If the evaluation function is well made then the deviation is significant in the change in balance in the game. The distances from the 
equilibrium forms a distribution. The greater the distance from equilibrium, the more likely the win. So we can describe the uncertainty on the final outcome as the entropy H(V) of the random variable representing the value of the position as returned by the evaluation function. The closer a value obtained during search to the
initial equilibrium the higher the uncertainty and the more distant, the lower the uncertainty. We may consider the size of distance resulted after one move
as a measure of informational gain. 
The information gain between position 2 , $p_2$ and position 1,  $p_1$ can be defined as

	\begin{equation}   I_{gain}(p_2,p_1) = H(p_2) - H(p_1)   \label{EQ}\end{equation}
 
Because the magnitude of the deviation from equilibrium signaled by an evaluation function must be a measure of the probability of the position having a 
certain absolute value, then

	\begin{equation} H(p_2) = f( k_1* v_2 ) \label{EQ}\end{equation}

and

	\begin{equation} H(p_1) = f(  k_2* v_1 ) \label{EQ}\end{equation}

It results

	\begin{equation}   I_{gain}(p_2,p_1) =   f( k_1*  v_2 ) - f( k_2*v_1 )   \label{EQ}\end{equation}

If the assumption of a linear dependence is made

	\begin{equation}   I_{gain}(p_2,p_1) =   k_1*  v_2  -  k2*v_1  + k_3   \label{EQ}\end{equation}

The conclusion is that moves which produce the highest variations in the evaluation function are the most significant assuming the evaluation function is ''reasonable good''. 
Such moves are captures of high value pieces. The entropy rate of a pieces is in general in a logarithmic relation with the mobility factor of that piece as it has been previously
shown.

In the above equations it is assumed the relation is linear, however it is possible to assume also a logarithmic relation for the relation between the 
material differences from the equilibrium and the uncertainty in regard to the perfect value of the position.Assuming the relation is described by a logarithmic function,

	\begin{equation} H(p_2) = \log{( k_1* v_2)}  \label{EQ}\end{equation}

and

	\begin{equation} H(p_1) = \log{(k_2* v_1)}  \label{EQ}\end{equation}

 and assuming $k_1$ = $k_2$ = 1 ,

	\begin{equation}   I_{gain}(p_2,p_1) = \log{v_2}  - \log{v_1}   \label{EQ}\end{equation}

	Approximating however like in the previous calculation for trajectories the position through the components and describing the entropies as the sum of the entropies of components ( pieces , subsystems) which makes sense for the above mentioned reasons the equations become

	\begin{equation} H(p_2) =\sum{ \log{ v_2}_i }  \label{EQ}\end{equation}

and

	\begin{equation} H(p_1) = \sum{\log{ v_1}_i}  \label{EQ}\end{equation}

which is a similar result to that obtained using the uncertainty on the position and trajectories of pieces. In one case it has been used the assumption that pieces follow a random walk trajectory and in the other case , the last calculation it has been assumed the relation between the entropy of a move is logarithmic with the distance from 
the matherial value equilibrium at the beginning of the game. 
This corroborates to confirm the approach and this is also confirmed by the experimental evidence in favor of the model.

	\subsection{A structural perspective on evaluation functions in computer chess}
	
The design of the evaluation functions in computer chess and other games is not based at this time on a mathematically proved method. 
The formula proposed by Shannon in  ~\cite{Shannon} is  

	\begin{equation}	Value(position) = \sum_{i=1}^{N}W_i  \label{EQ}\end{equation}

and the summation considers all the elements taken in consideration and observed on the board. The structure of the evaluation function  given by the above mentioned article written by Shannon has been the first such design and is
a simple one compared with modern engines using as much as 6000 elements in the evaluation function.

While in the formula published in the first paper on the topic    ~\cite{Shannon} the evaluation elements can be taken directly from the board and a program can do this with a high precision, for modern functions, we can assign a probability associated with the ability of the program to evaluate 
correctly a feature. One can number the pieces easily but it would not alway be so easy to detect more complicated positional and strategical patterns.
For the general chess NxN the evaluation would imply very complex patterns and will result in certain probability for the recognition of positional structures.
In this case there is the probability of a correct recognition for each feature of the evaluation function.This probability is the probability in the general formula that described the entropy.

	\begin{equation} H_{value} = - \sum_{x \in E_s} {p_{E_i}(x)*\log{ p_{E_i}(x)}}   \label{EQ}\end{equation}

where $E_i$ is the evaluation feature i and $P_{E_i}$ is the probability associated with the recognition of the feature.

This is the structural representation consistent with the uncertainty on the position and its modeling as entropy. This corroborates with previous facts in making certain a distribution and its 
associated entropy is the more general and correct way to describe the model.

	\subsection{The mutual information of positions in chess and the relation between the entropy of a node and the entropy of its descendants}
	
	Often the best moves and strategies in similar positions correlate and this is the assumption behind the theory of chess in its most important areas, strategy, tactics,
openings, end-games. In similar positions often similar strategies or tactics are used and openings and endgame moves are repeatedly used.

	It is possible to consider as the cause of the mutual information of two positions the number of evaluation elements $E_i$ having the same value 		for two positions. 

	Let X be a variable representing the value of a position and having a certain distribution and Y be a different variable representing the values of 
	a position near the first one. Then the mutual information is described by

	\begin{equation}  I(X_{v} ,Y_{v}  ) = \sum  p( v_{E_{i_x}} , v_{E_{i_y}} ) \log { \frac{ p( v_{E_{i_x}},v_{E_{i_y}}) } {  v_{E_{i_x}} v_{E_{i_y}} }   }    \label{EQ}\end{equation}

	where $v_{E_{i_x}}$ = $v_{E_{i_y}}$ or it is sufficiently close and  $p( v_{E_{i_x}},v_{E_{i_y}})$ is the probability that the two random variables representing analogous evaluation features applied to the two different positions take the same value.

	The information about a position considering the value of the siblings is

	\begin{equation}    I(X_{v} ,Y_{v}  ) = H(X_{v}) - H(X_{v}|Y_{v}) \label{EQ}\end{equation}

	We can consider each evaluation element as a stochastic function. When a piece is captured the number of stochastic functions will decrease
	and the relative entropy of the position will also decrease along such variation.

	Therefore a big change in one of the evaluation elements, such as a capture, will result in a big information gain. This is very much consistent with 
	the observations in computer chess where programmers place active moves to be searched first in the list of moves.

	The uncertainty of a position can be eliminated by exploring the nodes resulting from the position. This observation results in:

	\begin{equation}	H(position) = \sum H( descendant(position) )	 \label{EQ}\end{equation}

	The reality is that because neighboring positions have mutual information, the joint entropy is smaller than the summation of individual entropies
of the positions resulted from the original position. 
If we quantify the elements in the evaluation function, equal for pairs of nodes, a significant number of common elements are the same. This will result from the observation of a number of positions assuming an evaluation function given. However, it is clear that the position cannot differ by more than a
10\% value if no piece has been captured at the last move. And if such capture has been realized, then it is perfectly quantifiable. Rarely the strategic patterns which are harder to quantify could change after a move more than 10\% of the material value.
 
	The equation becomes

	\begin{equation}	H(position) \leq \sum H( descendant(position) )	 \label{EQ}\end{equation}

	\paragraph{Verification by the method of algorithmic information theory}

	We can verify this idea by using a model based on algorithmic information theory. It will result a new verification for the reasoning above.

	The  game tree is an and/or tree , where the program-size complexity H($p_1$,$p_2$,$p_3$,...) of the set of descendants of a node is bounded
	by the sum of individual complexities H($p_1$) , H($p_2$) , H($p_3$)...  of the descendant nodes. 

	\begin{equation}  H(p_1,p_2,p_3,...) \leq  H(p_1) + H(p_2) + H(p_3) + ... + C	\label{EQ}\end{equation} 

	The same expression may hold also for elements of an evaluation function.

	\subsection{The information gain in game tree search}

	The reduction in entropy after moving a piece can be interpreted as the information gain caused by a move.

	\begin{equation}	I_{gain} = H_{before move} - H{after move}  \label{EQ}\end{equation}

	\subsection{Problem formulation}

	In the light of the new description it is possible to reformulate the search problem in strategy games.
	The problem is to plan the search process minimizing the entropy on the value of the starting position considering limits in costs. The best case is when entropy, or uncertainty in the 			value of a position becomes 0 with an acceptable cost in search. This is feasible in chess and it happens every time when a successful combination is executed and results in mate or significant advantage. 

	It is possible to formulate the problem of search in computer chess and in other games as a problem of entropy minimization.

	\begin{equation} Min \{ H(position)  \} = Min \{ -\sum_{i=1}^{\infty}{P_i}log P_i \}  \label{EQ}\end{equation}

	subject to a limit in the number of position that can be explored.

	\section{Results}

	\subsection{Consequence 1: The active moves produce a decrease in the entropy }
	
	In chess, active moves are considered moves, such as captures, checks .  It results according to this model that,  such moves will cause a reduction in the entropy of the current position during the exploration of a variation with 
	\begin{equation}	log(weight Of The Piece) = log(K*number Of Moves) \label{EQ}\end{equation}

	Entropy rate is applicable to stochastic functions. It is possible to associate entropy to a set of stochastic functions. When the number of stochastic 	functions vary, also entropy will vary. This may be seen as the entropy rate of the system. 

	In this model, each piece can be seen as a stochastic function and the variation in the number of such functions will generate an entropy rate.

	Capturing a queen results in this system in a reduction with log(28) of the uncertainty of the position, capturing a rock results in the 
	decrease of uncertainty with log(14) , capturing a bishop results in the decrease of entropy with log(14). This is significant because the mobility is correlated practically with both uncertainty on the outcome of a position and with material gain. This fact is very intuitive.  The reduction of active pieces gives us also a
	measure of the reduction in the branching factor which causes a reduction in the complexity of exploring the subtree and a higher increase in accuracy for a certain 
cost of exploration. 

	This is very well seen in practice because such moves correlate with decisive moments in the game. There is good evidence for the fact that 
	exchanges and captures are orienting the game towards a position where the outcome is clear, where there are few uncertainties. This is true also in Shoji. Experimental evidence 
	used for the optimization a partial depth scheme using data from games confirms the conclusion obtained here in a different way. ~\cite{Tsuruoka}

	\subsection{Consequence 2: The combination lines as a cause of decrease in uncertainty on the value of the game }

	Because lines containing combinations often include many captures, according to the model described in this article such variations cause
	 a decrease in entropy of the position from where the variation starts and therefore cause a decrease in uncertainty about the game. This conclusion is also very well supported by observations, it has very good experimental verification. This is easy to test in a game and observe that combinations end with a clear position, mate or a decisive advantage  on one side and the uncertainty is 0 or very low.
	If the combination fails, often the side undertaking it will not be able to recover the lost pieces and would likely loose in such position and then the uncertainty is also near 0, because the outcome is clear. 

	It can also be observed the fall of the branching factor in the combinatorial lines and the fall in the number of material, resulting in an accelerated 
phase transition towards the end of the game. The number of responses from adversary is small during a forceful line resulting in less uncertainty in 
regard to adversary responses. Therefore it is no need to calculate all the responses. This is called initiative in chess.

	\subsection{Consequence 3: The information theoretic model and the information content of knowledge bases}

	The knowledge base can be understood as both a database or a knowledge base  of a human player. This is why the model described here unifies in a single theory the human decision making at the chess board as well as the computer decision making because reduction of uncertainty in a position by gaining information from the exploration of the state space is critical for decision making in both man and machine. In human chess it is called calculation of variations, in machine it is called search. The essence of gaining information in the information theoretic sense of the concept during the analysis of a position is the critical skill in human and machine decision making. 
	
	Let the probability of a trajectory (or move category) chosen in a position be,

	\begin{equation}   P_{trajectory} = \frac{N_c}{N_P} \label{EQ}\end{equation}

	where $N_c$ is the number of times the trajectory is chosen in the knowledge base and $N_p$ is is the number of cases the trajectory would have been possible.

 The knowledge base can reduce the uncertainty in terms of both moves from played games as well as combinations of categories of moves in a trajectory.There is a duality between the two perspectives and we may see the problem in both angles. The knowledge base can be used as  source of moves as well as a source of semantic representations and this happens also in the decision making of any human player. 

While the uncertainty of a string of moves finds its measure in the frequency of that variation being chosen if possible, the uncertainty of a trajectory finds its measure in the entropy of the string of symbols from the alphabet composed of move categories describing the semantic interpretation of a trajectory.

Often, there are correlations between the best decision in a position and the best decision in a different position, provided there is mutual information
between the two positions. The correlations are both at the level of moves as well as at the level of trajectories.

To certain trajectories can be associated probabilities according to the frequency of choices in a knowledge base relative to the number of times the trajectories have been possible. This creates a distribution and the uncertainty associated with it.  Let $Y_{knowledgeBase}$ be a random variable describing the trajectories from a knowledge base under the distribution given by the frequency of the decisions associated with the choice of trajectories . Let $X_{d}$ be a random variable describing the possible trajectories decided by a code along with the probabilities associated. 
 
The conditional entropy $H(X_{d} |Y_{KnowledgeBase}  )$ is the entropy of a random variable $x_{d}$ corresponding to the decision of the code, considering the knowledge of another random variable $Y_{KnowledgeBase}$ corresponding to the distribution of choices in the knowledge base. The reduction in 
uncertainty due to the knowledge of the other random variable can be defined as the mutual information of the two positions and of the associated tactical and strategic configurations. 

If we trust the knowledge base as resulting from games of strong players, then the uncertainty of the chess or other strategic system in taking a decision in a similar circumstance is smaller.  In the conditions when the two positions have similarities there must be a significant amount of mutual information between the 
two distributions, decreasing the uncertainty of decisions. 

The mutual information in regard to the choices in the knowledge base and the possible choices in a position where a decision must be taken is

\begin{equation}  I( X_{d} , Y_{KnowledgeBase} ) = H(X_{d}) - H(X_{d}|Y_{KnowledgeBase}) \label{EQ}\end{equation} 

and

\begin{equation}  I(X_{d},Y_{KnowledgeBase}) = \sum  p(x_{d},y_{KnowledgeBase}) \log { \frac{ p(x_{d},y_{KnowledgeBase}) } { p(x_{d}) p(y_{KnowledgeBase}) }   }    \label{EQ}\end{equation} 

where $p(x_{d})$ is the probability that $x_{d}$ is the right trajectory to analyze in our position or the right trajectory to choose,
 $p(y_{KnowledgeBase})$ is the probability that $y_{KnowledgeBase}$ is chosen in the database record (and we assume this is also the probability that the decision is good)
and $p(x_{d},y_{KnowledgeBase})$ is the probability that both are right strategies in each respective position. 
As it may be seen, $p(x_{d},y_{KnowledgeBase})$ depends on the tactical and strategical similarity of the two positions given by the mutual information of the two positions.

The value of information can be measured experimentally in the increase of decision power in chess programs resulted after the addition of knowledge bases.The knowledge base can refer to opening database, endgame database, and knowledge for orienting exploration.
The addition of knowledge to a program is also a particular case of those previously mentioned, because the theory of chess is resulted from the analysis of games. 
The effect of theory addition on a programs power is known and has been measured by  ~\cite{JS}. The increase of decision power by the addition of endgame and opening
bases has been impressive. The measure is at this time specific to the application and has limited generality as long as a general system architecture for such programs is not
defined in a mathematical way. Theory is the practice of masters and therefore the above mentioned relations explain the increase in power in programs after using chess theory by correlating the moves with those of the masters who first introduced the theory through their games. 
The increase in program power with the addition of knowledge can be used to measure the mutual information of positions. The experiment is clearly possible and the result is very much predictable, $I( X_{d} , Y_{KnowledgeBase} )$ is something dependent on the knowledge base and the heuristics used by the program. It can be used as a measure of performance by people developing programs.
This is the mathematical explanation for the increase in performance when a program uses knowledge bases. For the particular case when the knowledge base contains 
perfect values for endgame, and the positions are the same it is obtained as expected the uncertainty or entropy of the decision is 0. The reduction in entropy is based on the
size of the endgame table-base which is a measure of the kolomogorov complexity of the position if the endgame base is optimally compressed. So the reduction in computation
time is a trade-off with the size of the code, including the knowledge base size.
For other cases when these particular conditions are not met such approach reduces the uncertainty in selecting a line of search for analysis but the entropy does not become 0. This explains also the advantage of knowledge in human 
players and it may possibly explain also the formation of decision skills in humans.  The decrease in entropy gives a measure of the quality of information we have from the knowledge base.   The assumption is that positions have  mutual information which is sometimes verified. The tactical and strategical patterns
may be described as positions with a significant amount pf mutual information and known trajectories of play where the probability of a certain  outcome $p(y_{KnowledgeBase})$ is statistically significant.

For the search process it represents an information gain because it is possible to be more certain about the outcome in this way. It is true, for the particular case of 8x8 chess the idea of analysis of the 
problem of strategy and tactics using the mutual information of correlated subsystems may seem an abstraction. However this is likely to 
represent a foundation of the theory of strategy and tactics for the general problem of NxN chess and with this to connect the problem to the other important problems in computer science and science. Controlling the game may be formulated as a problem of controlling these systems. This represents
a generalization of the concepts of tactics and strategy. Strategic and tactical plans may be seen as particular cases of optimal control policies where the control policy is based on uncertainty reduction. A system in this model, controls the game by controlling the options offered to the adversary.

	\subsection{Consequence 4: The correlations between decision-making in different games of a player}

There is a second way to use mutual information. Instead of referring to the database we could compare the previous games of a player and the 
choices he made in previous games to predict and anticipate the choices from a future game. Studying opponents games is very important for players
at a certain level. One can investigate what would most likely the opponents  play before the real game has taken place. 
The previous equations can be used in this conditions.

  $I( X_{ADVERSARY} , Y_{PreviousAdversaryDecision} )$ = 

\begin{equation} =  H(X_{ADVERSARY}) - H(X_{ADVERSARY}|Y_{PreviousAdversaryDecision}) \label{EQ}\end{equation} 

and

  $I(X_{ADVERSARY},Y_{PreviousAdversaryDecision})$ =

 $= \sum  p(x_{ADVERSARY},y_{PreviousAdversaryDecision})$ *

\begin{equation}*\log { \frac{ p(x_{ADVERSARY},y_{PreviousAdversaryDecision}) } { p(x_{ADVERSARY}) p(y_{PreviousAdversaryDecision}) }   }    \label{EQ}\end{equation} 

where $p(x_{ADVERSARY})$ is the probability that $x_{ADVERSARY}$ would be the trajectory chosen by the adversary in this position ,
 $p(y_{PreviousAdversaryDecision})$ is the probability that $y_{PreviousAdversaryDecision}$ has been chosen in the previous games of the adversary
and $p( x_{ADVERSARY} , y_{PreviousAdversaryDecision})$ is the probability that both would be chosen in similar positions. \newline
$p( x_{ADVERSARY} , y_{PreviousAdversaryDecision})$  depends on the tactical and strategical similarity of the two positions given by the mutual information of the two positions and the predictability of the adversary. From this it results that it pays of to be less predictable in choices such that the adversary is uncertain and does not know
what to prepare in defense. This is why the randomization of strategies plays a critical importance in human and computer chess. Not using information theory but by a practical design idea Shannon suggested a statistical variable to be left in a chess computer so that the opponent cannot follow always a certain winning path after he found it.

	\subsection{Consequence 5: The problem of pathology and its information theoretical explanation}
	
	The model presented predicts a decrease in the entropy of the search trajectory and in the uncertainty on the positional value on the lines with traps and combinations which happens in reality. One can see for example the experiment with the position presented. The evaluation is perfect , the value of the position is 1, win.
	This gives also a good explanation why chess is not a pathological game as defined by Dana Nau in ~\cite{DSNau1} ~\cite{DSNau2}  ~\cite{DSNau3}  ~\cite{DSNau4} ~\cite{DSNau5}. 
	The model described here offers a theoretical explanation for the unquestionable evidence from chess and computer chess as well as for the 
	explanation of J. Pearl in regard to why chess is not a pathological game. 

	 For instance, in the example, after the execution of the search, the uncertainty is 0 because the search
proves the possibility to force mate, regardless of the response of the adversary, considering the moves are legal according to the rules of chess.

	The model described here explains in a more general way the causes of pathologic search. A pathologic search process can be defined
	as a search process where the uncertainty on the value of the position increases with the search depth. It can easily be seen that a search process where the 
	information gains per search ply is below a critical value will be pathological. The rate at which the heuristic can obtain information by exploring
	the state space depends on its ability to extract information as well as the general characteristics of the state space.

	The equation that gives the decrease in the entropy on the position from where the search is executed with depth is the equation that relates the entropy of a parent node to that of children nodes.
 
	Let $p_1$, $p_2$,... positions resulted from a node by application of the possible moves. Then
	
	\begin{equation}  H(p_1,p_2,p_3,...) \leq  H(p_1) + H(p_2) + H(p_3) +....+ C	\label{EQ}\end{equation} 

	That means the joint entropy of the positions resulted from a node is smaller than the summation of their entropies.  The summation of their
	entropies can be considered as shown above to be approximatively equal to that of the parent node . If the process is continued to infinity and the previous
	condition is true for each level or at least it describes a trend then the entropy will decrease to 0 at some point provided that the rate of decrease level by level
	is not infinitely close to 0. It can be conjectured this is the explanation for the increase in power strength of good chess programs with a greater depth of search.
	It can also be tested experimentally on various functions and probably evidence of this phenomenon will emerge. A full proof is a future objective.
	Any proof must take in consideration a mathematical description of the optimal chess program. This is more than mathematics can handle at this time.

	The higher the value of the constant C, the less uncertainty will be about the value of the initial position, for each search ply.

	It is not necessarily that C is a constant, it can be also a variable dependent on the search level and path. As the previous example shows,
	on combinatorial paths, C(depth) is higher as the search procedure gains information at a higher rate and the joint entropy of the descendants
	of a node is significantly smaller than the entropy of the parent node. Both mutual information as well as information gain in transition 
	from a node to its children explain this.

	The smaller branching factor in the combinatorial lines is due to exchanges in pieces. The exchanges in 
	pieces alone may cause a smaller branching factor. The checks reduce the options even more.
	It results that the combinatorial lines where the time of search is often minimal for a certain increase in accuracy and the probability of a good
	solution is higher than for other variations, must have a high priority.The greater amount of information is accumulated fast, a greater number of later 
	decisions may use more information and overall a higher proportion of decisions taken during exploration will benefit from the early gain in
	information. This explains why detecting a mate or a very good solution fast is so effective in decreasing the search time in a combination.
	This will be seen from the experiments performed for this article.	
If a search knows it can force mate than it has optimal information and can discard any other line.

	This depends in practice on the type of position as well as on the way the evaluation function is designed. Any experiment is highly dependent
	on the design of the function. The existence of correlations between close nodes has been considered by Dana Nau a reason explaining why chess is not
	pathological. An alternative explanation, according to the information theoretic model of computer chess, developed here can be give. The mutual information of the
	siblings  explains the correlations of their values. The cause of this correlation has been shown previously in this paper.
In addition, if any error occurs in the elements of the evaluation function responsible for mutual information then the error would not effect the 
	quality of the decision. Experimental studies by Beal and Ivan Bratko have shown the existence of correlation between siblings. This is not only a confirmation of
the hypothesis of Nau in regard to his theory but also a confirmation of the model constructed in this paper which has a completely different foundation and meaning than that of Nau. Some conclusions of Nau as well as many more other previous important results from the core of computer chess,
 more than 20 can be seen in a different way and in some cases mathematically derived from the model described in this paper.
The correlation is a different 
	name for mutual information. Therefore, the mutual information between close nodes $p_1$ and $p_2$, I($p_1$,$p_2$)  explains the correlations 			between the values of close positions measured by the the previous mentioned authors. ~\cite{IB1} ~\cite{IB2} ~\cite{IB3} ~\cite{IB4}       			~\cite{IB5}

	\subsection{Consequence 6: The value of pieces and their information content }

	Because the mobility factor, there is a mathematical connexion between the value of pieces in various systems and the relative entropy of a piece as 					described here. The relative entropy of a chess piece as well as the value of a piece are both a function of mobility but in a different way.

	\includegraphics[width = 2.5in , height = 2.5in]{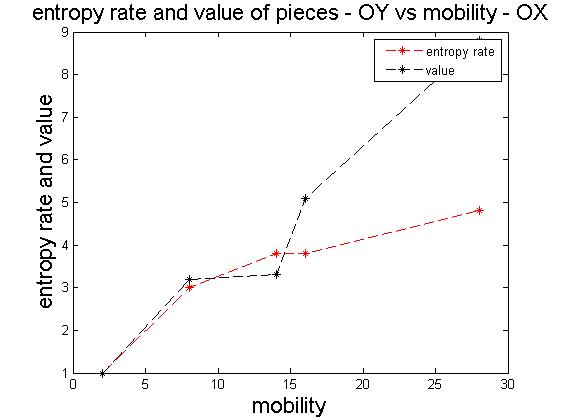}

	Usually the values of pieces correlates with mobility.
	The only exception appears to be the values of knight and bishop which have similar values despite a significant difference in mobility. The bishop
	is usually given a better score for mobility but not that much better compared to the score of the knight as it is the difference in mobility.
	This difference between theory and practice can be solved easily if we consider instead of mobility, a random walk or a trajectory in terms of chess.
	If in terms of mobility, the mobility of a bishop is much higher , then in terms of trajectory, the usual trajectory of a bishop which is the measured by 
	a possible random walk is not much longer or better than that of a knight simply because the knight in the center of the board will always have 8
	moves while the bishop will rarely have 14 before the endgame. So it is their entropy that is much more closer to their value as pieces that the 
	mobility.  If somebody has seen enough positions it is clear so. If not it is easy to imagine an experiment calculating the real number of moves and
	the length of a trajectory, the random walk possible without taking positions in fields defended 
	by opposing pawns. Also the trajectory of the knight may allow this piece to infiltrate behind defenders. So it is the trajectory a measure of 
	power more than mobility.

	The same explanation can be given for the big difference in value between bishop and rock despite the small difference in mobility: the rock operates
	on columns and the trajectory makes it possible a battery of two mutually defending rocks which is stronger usually than 2 bishops. The two rocks
	can attack common objectives and therefore the trajectory is the measure of value to higher extent than the mobility. This justifies the approach
	of using trajectories for the model presented. 

	\subsection{Consequence 7: The ordering of moves and the entropy reduction in the variations}

	It is a consequence of the previous analysis that lines with many active moves such as captures and combinations lead to less entropy in the end position and a more accurate evaluation faster than other lines of search. So the best ordering, according to the entropy reduction search is when 
the moves resulting or anticipating a high reduction in entropy and a high gain in information must be prioritized first in the search, in order to take the next exploration 
decisions in a more informed way. This conclusion may be verified with logic.

	Consider the best order of search (in general not in computer chess) given by Solomonoff at ~\cite{Solomonoff}

	\begin{equation}   Order = \frac{T_i}{P_i}  \label{EQ}\end{equation}

	where $T_i$ is the time needed to test the trial, and $P_i$ is the probability of success for that trial.

	For combinations $T_i$ is often the shortest and $P_i$ is the highest because there are relatively few combinatorial lines compared to the amount of variations but combinatorial lines account for a disproportionate percentage of matting and winning lines. Shannon called these lines high probability
lines, because of course it is much more likely that such lines will result in a change in the equilibrium and may lead to a new equilibrium.  Here it has been
shown using two methods ,the information theoretic method and through logic why. 
	
	The experimental verification of this conclusion is clear in that almost all championship programs calculate the active line first based on experimental
evidence. Also this explains why chess books recommend the investigation of tactical lines first.
	
	\subsection{Consequence 8: The  game theoretic and information theoretic model}

	It may be seen that often end-games are more likely when there are fewer pieces on the board, the game is often closer to the objective
           in such state. In these circumstance the evaluation is more
	accurate in such positions. In this model the game theory is combined with information theory such that the information as a concept in 
	game theory is expressed through an information theoretical concept.

	\subsection{Consequence 9: The quiescence search and the entropy }

	The quiescent search represents the ultimate reduction of entropy. Actually the heuristic has been intended for realizing a state of equilibrium
	so that static evaluation functions can be used. This confirms the prediction of the model described here in regard to the decrease in entropy and 
	uncertainty by captures, checks and other moves. Actually the definition and the model of search based on entropy, especially in its physical 			meaning can justify also the name of the quiescent search method in a formal way and it can give its definition.
	
	\subsection{Consequence 10:  A definition of information gain in computer chess}

	It is possible to define the information gain during the search process based on the reduction in uncertainty in the following way:

	\begin{equation}    I_{gain}  =  \triangle H         \label{EQ}\end{equation}

	Where H represents the uncertainty in the value of the position and  $\triangle H$    

	\begin{equation}	\triangle H = H_2 - H_1   \label{EQ}\end{equation}

	represents the variation of uncertainty in the current position after a move is made. It is the information gained after making a move.

	In the case when \begin{equation}  \triangle H  \le 0 \label{EQ}\end{equation}  we speak of information gain, 

	if \begin{equation} \triangle H \ge 0 \label{EQ}\end{equation}  we understand information lost through approximate evaluation or other operation.

It is possible to describe the information gain of the search process by defining the heuristic efficiency

	\begin{equation}	HE =  \frac{I_{gain}}{\triangle Nodes}  = \frac{\triangle H}{\triangle Nodes}      \label{EQ}\end{equation}

	When   $\triangle  Nodes$   $\longrightarrow$ 1 the information gain results after a move is

	\begin{equation}    I_{gain}(Move) =  H(before Move) - H(after Move)         \label{EQ}\end{equation}

	This concept may be considered similar to the the concept of information gain for decision trees, the Kullback-Leibler divergence.
	We may see the same principle also here, the higher the difference between entropies, the higher the information gain, which makes very 			much sense also intuitively and it provides a new theoretical justification for the empirical heuristics of chess and computer chess. 

	\subsection{Consequence 11:  Comparing heuristics based on the efficiency in the reduction of uncertainty in search}

	It is possible to define the domination of a heuristic exploration strategy H1 over other heuristic H2 over a certain number of nodes if for that number of nodes the first heuristic H1 decreases the uncertainty on the evaluation of the position more than H2.
	A heuristic H1 dominates other heuristic H2 if for every input instance, the information efficiency of one of the heuristics $HE_{1}$   $\leq$  $HE_{2}$.	
	It is very intuitive to understand that an exploration strategy providing more information about the state space will result in a better decision and one player
or decision-maker will dominate.

	\subsection{Consequence 12:  The justification of heuristic information with the information theoretical model of search}

	Sometimes, heuristic information has been considered only for the case of the upper and lower limits of the value of a solution. It has been 				recognized also in the past that such limits would decrease the search time. However the previous description of heuristic information did not 
	take in account the nature of the moves. It is natural to believe that moves themselves provide the bounding for the estimation of the value of a positions, so the
	quantification of the position as described here has a connexion to the methods which quantify the upper and lower bound only. The model is however more general 
because assumes a distribution , not only an interval.
 It has been shown
	experimentally that certain moves  lead to a certain expectations also for other similar games ~\cite{Tsuruoka}.

	\subsection{Consequence 13:  The decrease in uncertainty caused by a search method and the domination on the chess board}

	It can be seen that a decrease of uncertainty is not an absolute measure but it is the view of one of the players depending on the investigation
	methods used. A method achieving better information will have better chance of detecting the best strategy and of winning.

	\subsection{Consequence 14 : The justification of the partial depths scheme using the information theoretic model.}

	The partial depths method is a generalization of the classic alpha beta in that it offers a greater importance to moves considered significant
	for the search. If all moves have the same importance then , the partial depth scheme can be reduced to the ordinary alpha-beta scheme.
           It can be described also as an importance sampling search. The partial depth scheme has been used by various authors. As Hans
	Berliner observed, few has been published about this method   ~\cite{HBerliner2}.The contribution of this article goes in this direction.

	It is possible to define a function returning the depth:

	\begin{equation}	\triangle depth = f( path )   \label{EQ}\end{equation}

	This is a generalization of the classic alpha-beta because in classic alpha-beta $\triangle$ depth $=$ constant;
	If the decision to add a certain depth to the path is dependent only on the current move and position , 
then if $\triangle$ path ${\longrightarrow}$ 1
	the decision depends only on the current position.

	The increase in depth is dependent on the path in this method, where the path is composed of moves  $m_1$ ,$m_2$ , $m_3$ , .... .  In the classic alpha-beta the depth increase is constant regardless of the type of move.
	
	\subsection{Consequence 15: The design of a search method with optimal cost for a certain information gain}

	The principle behind a theory of optimal search should be the allocation of search resources based on the optimality of information gain per cost.
	It results that the fraction of a search ply added to the depth of the path with a move should be in inverse proportion to the quality of the move.
	The standard approach gives equal importance to all moves, the fraction ply method gives more importance to significant moves.
	Therefore it must be described a quantitative measure for the quality of a move. The reduction from the normal depth of 1 ply should be 
	proportional to the quantitative measure of the quality of a move.

	The fraction ply FD must decrease with the quality of the move relative to optimal. The fraction ply added would be equal in this system to the decrease of a full play with the approximate entropy reduction achieved by that move compared
	to a move having the highest entropy reduction.

	For instance for a capture of a rock the entropy reduction is $\log 14$ 

	\begin{axiom}
	
	An axiom of efficient search in chess , in computer chess and in search in general should be that the probability of executing a move must be equal to the heuristic efficiency of that move which is equal to the information efficiency of expanding the node resulted after the move. The same principle can be considered in general for trajectories.

	\end{axiom}

	By notation, let the heuristic efficiency be HE and   $P_{c_i}$ be the probability of a move in category   $c_i$ to be executed. 
	The heuristic efficiency is a fundamental measure of the ability of a search procedure to gain information from the state space. The heuristic efficiency depends in this
	analysis on the categories of moves and trajectories defined. The examples are for moves with individual tactical values, however the analysis can be extended also
	to tactical plans generated by pins, forks and other tactical patterns. Because such analysis would require some readers to look for the meaning of these structures in 			chess books and also because space considerations the moves generating such configurations would not be presented as examples. However, no additional theoretical      	difficulties would emerge from the introduction of these move categories. The same applies to strategical elements. 

	Following the principles outlined, a formula for the fraction ply can be derived.

	\begin{equation}	P_{c_i} = k* HE  \label{EQ}\end{equation}

	Considering that 

	\begin{equation}   HE = \frac{  I_{gain_i} }  { \triangle Cost  }  \label{EQ}\end{equation}

	and

	\begin{equation} \frac{ I_{gain_i} }{ \triangle Cost } = \frac {\triangle Entropy_{category_i } } { \triangle Cost  }    \label{EQ}\end{equation}

	it means 

	\begin{equation} HE = \frac{\triangle Entropy_{category_i} } { \triangle Cost  } \label{EQ}\end{equation}

	For k = 1, \begin{equation}  P_{c_i} =  \frac{\triangle Entropy_{category_i} } { \triangle Cost  } \label{EQ}\end{equation}

	from this, \begin{equation}   \triangle Cost = \frac{\triangle Entropy_{category_i} } { P_{c_i}  }  \label{EQ}\end{equation}

	Of course a different value than 1 can be given to the constant k and this will propagate without changing the meaning of the equations. The constant k would increase the flexibility of implementations actually, offering more freedom in this direction. 
	Now consider the same equation for the move category with the best information gain.

	It means \begin{equation}  P_{c_{BestGain} } =  \frac{\triangle Entropy_{BestGain} } { \triangle Cost  } \label{EQ}\end{equation}

	Assuming the moves from the best category, the most informational efficient will always be executed in the search, the following condition must be satisfied: 

          \begin{equation}  P_{c_{BestGain} } = 1 \label{EQ}\end{equation}

	Then  	\begin{equation}	\frac{\triangle Entropy_{BestGain} } { \triangle Cost  } = 1   \label{EQ}\end{equation}

	so

	\begin{equation}  \triangle Cost = \triangle Entropy_{BestGain} \label{EQ}\end{equation}
	
	The cost for execution of any of the two moves is the same. Equating this cost, it results

	\begin{equation} \frac{\triangle Entropy_{category_i} } { P_{c_i}  } = \triangle Entropy_{BestGain}  \label{EQ}\end{equation}

	It means 

	\begin{equation}  P_{c_i} = 	\frac{\triangle Entropy_{category_i} } { \triangle Entropy_{BestGain} } \label{EQ}\end{equation}

	which is a very intuitive result.

	In general, for a $trajectory_i$, the probability of a trajectory to be explored should be in this system
	
	\begin{equation}  P_{trajectory_i} = 	\frac{\triangle Entropy_{trajectory_i} } { \triangle Entropy_{BestTrajectory} } * \frac{\triangle Cost_{BestTrajectory}}{\triangle Cost_{trajectory_i}} \label{EQ}\end{equation}

	\subsection{Consequence 16: The ERS* , the entropy reduction search in computer chess}

	Let $ P_{c_i}$  be the probability that a move is executed and one more ply is added to the search. 

	The size of the ply added should be function of this probability. It is logically to consider the size of the play as a quantity increasing with the
	probability of the move not being executed. The probability of the move not being executed is $1 - P_{c_i}$ therefore
	assuming an abstraction, a linear relation of the form:\newline size of ply = k*( probability of  a move not being executed )
	then the relation between the size of the ply and the probability of the move to be chosen would be for k = 1

	         \begin{equation}   D = 1 - \frac{\triangle Entropy_{category_i} } { \triangle Entropy_{BestGain} }   \label{EQ}\end{equation}

	This may be considered even a theorem describing the size of the fraction ply in computer chess and even for other EXPTIME problems under the above assumptions and resulting from the above calculations.

	Starting from the previous equation, it is possible to use the the relative entropies of pieces and positional patterns to implement the previous 			formula.

	 Consider the check as the move with the ultimate decrease in entropy because its forceful nature and because it has a higher frequency in the vicinity of the objective, the mate than any other move.  Then all the other moves may be rated as function of the check move. Let such value be $\log 30$ . However, here can be used a constant 
reflecting the above mentioned properties of such move. It must be noted that not all checks are equally significant. Several categories of checks can be introduced instead of 
a single check category. Also in the application, not all checks are equally important, check and capture for example gains a better priority but in this example not a smaller depth. 
		
	As a consequence, if the normal increase in search depth is counted as 1 for moves without significance the fractional ply for a check is:

	\begin{equation}  D = 1  - \frac {\log 30} {\log 30}	\label{EQ}\end{equation}

	then D = 0 in this system because the best move should be always executed and then the depth added should be 0.

	For a capture of queen the entropy rate of the system decreases with $\log 28$.

	Then the fractional ply for a queen capture is 

	\begin{equation}  D = 1  - \frac {\log 28} {\log 30}	\label{EQ}\end{equation}
	
	after calculations, D = 0.02

	For a capture of rock the entropy rate of the system decreases with $\log 14$.

	Then the fractional ply for a rock capture is

	\begin{equation}  D = 1  - \frac {\log 14} {\log 30}	\label{EQ}\end{equation}
	
	after calculations, D = 1 - 0.776 = 0.223

	Instead of using the entropy rates for calculating the size of the fractional depth it is possible to use the value of pieces which is strongly correlated
	for most of the systems with the entropy rate of the pieces.As it can be seen from the calculation above, the higher the differences in entropy
	between consecutive positions in a variation, the higher the information gain. This can be understood as a divergence between distributions
	of consecutive moves. The more they diverge the higher the information gain after a move.

	\subsection{ Experimental strategy }

	 Such information theoretical analysis has not been undertaken yet in this general way and to such extent for the partial depths scheme and for the other consequences for search and for so many areas of computer chess. The experimental verification of the theory outlined is not restricted to the algorithm and to the procedure implemented. The consequences are consistent to the knowledge accumulated in chess and in computer chess. The experimental search in this paper concentrates on the new heuristic presented. To some extent the 
	experimental results are dependent on the search algorithm. In order to isolate the effects and the consequences of the theory outlined, the search function presented contains no specific chess knowledge and no other search heuristic.
	Many test cases have been used but it is not the space for all in this context. The analysis shown here will concentrate on a single combinatorial position and investigate
	the effect of changing the parameters and especially the change of the value of the parameter that describes the reduction in the depth 				of a ply for significant moves. In part the theoretical
	considerations resulted from the information theoretic model have as consequences search principles similar to those used by chess masters but the 
	implementation does not follow up chess knowledge but shows how a similar effect can be obtained based on the theory. The success of some 
	principles of search used by chess masters and observed empirically by great chess players through the ages find a theoretical justification in 
	the theory of search and decision in computer chess presented here. The results are telling in regard to the power of the method compared to 
	a standard alpha-beta search. The method is a generalization of standard alpha-beta search because for the case when the depth added at each 
	play is the same we obtain the standard alpha-beta, with all the performance decrease. The experimental setup does not contain a quiescent 
	search or extensions at the limit of the search depth. No null move heuristic or any other heuristic is used because this would improve the 
	method and the effects would be seen as not necessarily that of the method proposed here but a result of a combination of heuristics which is of
	course not the case in our experiment. For a standard alpha-beta without any heuristic it is probably required a significant search power to find a 			mate as deep as 14 plies, while the method shown here finds it in something like 20000 nodes , very few by any standard. The position does
	not favor and is not particularly favoring the method proposed. Even without any additional heuristic the skeleton of the method solves 
	many of the combinations of this depth and even deeper. The test position is shown above at the explanation on what is a combination.
	The position represents the starting position of a combination 14 plies deep. It is not the deepest combination, however the depth is chosen 
	because only some of the most powerful chess machines,for example Deep Blue would search uniformly at that depth. Of course Deep Blue had many heuristics and 
	chess knowledge incorporated and did not rely only on search power. Deep Blue could have certainly solved even some of the deepest combinations, but
	on uniform search 14 plies is very much for a position with branching factor 50 at start. 

	The experimental strategy consists in changing the partial depth and observing the effect on the number of nodes expanded, on the maximum
	depth attained by the search, on the ability of the algorithm to solve the problem, on the distribution of depths of variations in the search process.
	It will be seen also what is the effect of the second search parameter, the maximum depth of the uniform search line on the ability of the program to 	solve the problem.
	  The metric for the amount of resource used , the ''cost'' of search is the number of nodes expanded in this experiment. The evaluation function is very fast,
	based only on material value and there are no researches of nodes in this schematic example. The number of nodes expanded shows the relative performance
	of the parameters used for the same algorithm. In addition, unlike nodes, the time would be to a great extent dependent on CPU and implementation, not on the algorithm. The number of positions expanded in this experiments is very small compared to what can be achieved in terms of nodes searched, even for PC-based applications.
	The number of expanded positions is comparable to what the computers of previous decades could search. 
	The use of a single and very general search procedure without other heuristics gives the measure of the performance of the formula derived in the article without making possible the interpretation that it works because other methods.

	\subsection{ Experimental results }

	The meaning of the columns is the following:\\
	column 1:EXPERIMENT NUMBER  - represents the number of the search experiment \\
	column 2:NODES SEARCHED  - represents the number of nodes searched in the experiment  \\
	column 3:TERM DIVIDING THE REDUCTION IN PLY - represents the number dividing the term decreasing the size of the normal ply added to the current depth \\ 
           column 4: MAX DEPTH ATTAINED - the maximum depth in standard plies attained , here it is added 1 for each ply \\
	column 5: MAX UNIFORM DEPTH - the maximum allowed depth in the partial depth scheme considering a step of 6 decreased with a value depending to the quality of the 		move  \\
	 column 6: SOLVED OR NOT - 1 if the case has been solved with the parameters from the other columns\\
	 column 7: STEP SIZE - the  number added to the partial depth for each new level of search in case of moves without importance  \\
	The following is the table with the results of the search experiments:
	\begin{center}
		\begin{tabular}{ | l  |  l  |  l  |  l  |  l |  l  |  l  |  l  |   }
			\hline

                            column 1 & column 2 & column 3 & column 4 & column 5 & column 6 & column 7  \\ \hline
		              1       &    20827   &   1              &  17              &     16           &    1           &            6        \\ \hline  
		              2      &     1080           &   1.25              &   8          &   16             &   0            &     6          \\ \hline 
		              3       &     88532           &   1.25              &    12            &   22             &     0          &    6             \\ \hline 
		              4      &      139545          &    1.25             &       12         &   24             &    0           &       6          \\ \hline 
		              5       &      155130          &     1.25            &   14             &   26             &    0           &       6         \\ \hline 
		              6       &   291714       &   1.25                &    14              &  28                &  0             &     6              \\ \hline 
		               7      &      82208          &   1.25              &      16          &      30          &   1            &   6            \\ \hline 
		               8      &   311166             &  1.5               &      12          &     32           &    0           &             6     \\ \hline 
		               9      &     494560           &   1.5              &      13          &     34           &      0         &              6   \\ \hline 
		               10      &   1009407             &  1.5                &   14                &    36            &    0           &            6      \\ \hline 
		                11     &   208423             &  1.5               &     15           &      38          &     1          &             6    \\ \hline 
		               12      &    1821489            &  1.75               &     13           &   40             &    0           &           6      \\ \hline 
		                 13    &     2337740           &   1.75              &    14            &    42            &    0           &           6       \\ \hline 
			                14   &     381146           &          1.75       &     14           &      44          &     1          &           6      \\ \hline 
		                  15   &     4547933           &  2.00      &     14           &     46           &          0     &           6      \\ \hline 
		                   16  &      603499   &     2.00      &     14       &   48     &      1    &           6        \\ \hline 
		                   17  &     8549650    &        2.25         &    14    &   50    &       0   &           6       \\ \hline 	
			\hline
		\end{tabular}
	\end{center}

	\begin{center}
		\begin{tabular}{ | l  |  l  |  l  |  l  |  l |  l  |  l  |  l  |   }
			\hline
		      
                            column 1 & column 2 & column 3 & column 4 & column 5 & column 6 & column 7  \\ \hline
		                   18  &      816524    &       2.25          &      14     &      52      &      1      &           6      \\ \hline 
		                   19  &         822539  &        2.5           &      14     &    54     &        1       &           6       \\ \hline 
		                    20 &     880194    &      2.75        &      14      &      56       &      1         &            6      \\ \hline 
		                    21 &      897504    &       3    &       14         &   58             &     1          &            6       \\ \hline 
		                    22 &        1026531   &      3.25           &      14          &    60            &     1          &            6      \\ \hline 
		                    23 &        2280040  &  3.5      &    14            &       62         &   1            &            6      \\ \hline 
		                    24 &       96973328    &        3.75         &      14          &    62            &    0           &            6      \\ \hline 
		                    25 &        3210105  &       3.75       &        14     &       64      &      1      &            6      \\ \hline 
		                    26 &         2661590  &      4.00           &     14           &       64         &      1         &            6    \\ \hline 
		                    27 &      4084892     &        4.25         &     14           &      66          &      1         &            6      \\ \hline 
		                    28 &       6624146   &      4.5           &     14           &    68            &       1        &            6     \\ \hline 
		                    29 &           4572359     &   4.75             &   14             &  69              &       1        &            6      \\ \hline 
		                    30 &        7711638    &   5            &    14            &     70          &      1    &   6      \\ \hline

			\hline
		\end{tabular}
	\end{center}

	\subsection{ Interpretation of the experimental results }

	\subsubsection{ (\romannumeral 1)   The step of the search  }

	At first a step representing a fraction of 1 has been used. However, better results have been obtained by using a step bigger than 1 for 
	not so interesting moves. The cause is the decrease in the sensitivity of the output and of other search dependent parameters in regard to the variations of other parameters and of the positional configurations.

	\subsubsection{  (\romannumeral 2)  The importance of detecting decisive moves early  }

	The detection of the variation leading to the objective early decreases the number of nodes searched very much. The fact that the mate has been found at 13 plies depth
	after only 20000 nodes searched shows the line to mate has been one of the first lines tried at each level, even without using knowledge. As it can be seen from the
	table if the mate is detected relatively fast the number of nodes searched is more than 10 times smaller.
	The next plot shows this. The maximums in the number of nodes represents the configurations ( a set of parameters ) for which the mate has not been fast detected.
	The OX represents the number dividing 	the factor giving importance to some significant moves and on OY it is represented the number of nodes searched. \newline

	 Plot of the increase in number of nodes when the importance given to moves with high information gain is decreased \newline         
	
	\includegraphics[width = 2.5in , height = 2.5in]{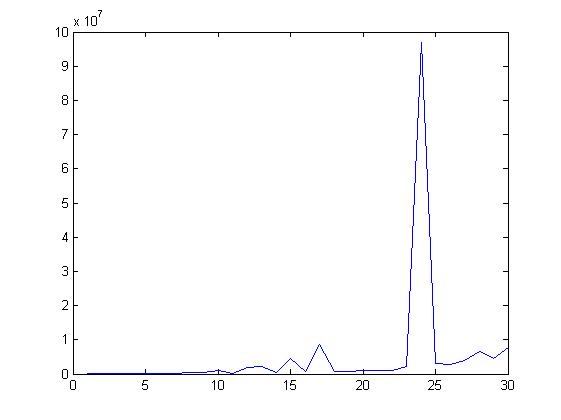} \newline

	On OX it is represented the virtual depth. On OY it is represented the number of nodes.
	
	As it can be seen, even a deeper search that detects the decisive line will explore less nodes than a shallower search that does not find the decisive line.
	For this heuristic and for most of the combinations, when the mate or a strongly dominant line is found fast, the drop in the number of nodes searched is as high 
	as 10 times, even if the uniform search is parametrized for a higher depth.

	\subsubsection{   (\romannumeral 3) The effect of the importance given to high information lines  }
	The number of nodes to be searched increases very much with the decrease of importance given to important moves and to lines of high informational value. 
	
	The following plot, based on data from the previous table shows the increase in the number of nodes explored with the decrease in the importance given to information
	gain when the solution is found. The less importance to the information gaining moves and lines is given, the greater the need for a higher amount of nodes to be searched in order to find the 
	solution.
	
	\includegraphics[width = 2.5in , height = 2.5in]{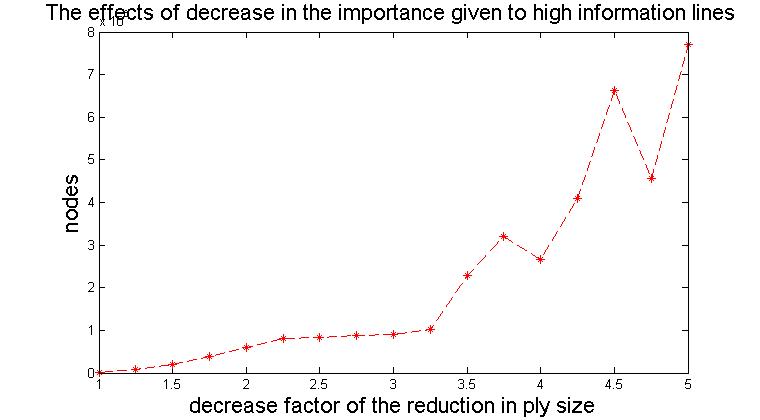}	\newline

	On OX it is represented the TERM DIVIDING THE REDUCTION IN PLY which represents the number dividing the term decreasing the size of the normal ply added to the current 		depth. On OY it is represented the number of nodes.
           The plot shows the explosion of nodes required to find a solution when the importance given to high information lines is decreased. As the importance given to 
	high information lines is decreased the number of nodes searched has to be increased. The importance given to information is decreased so the depth of search must 
	be increased to find the solution. 

	The following plot has the same significance but for the case when the solution is not found. \newline

	\includegraphics[width = 2.5in , height = 2.5in]{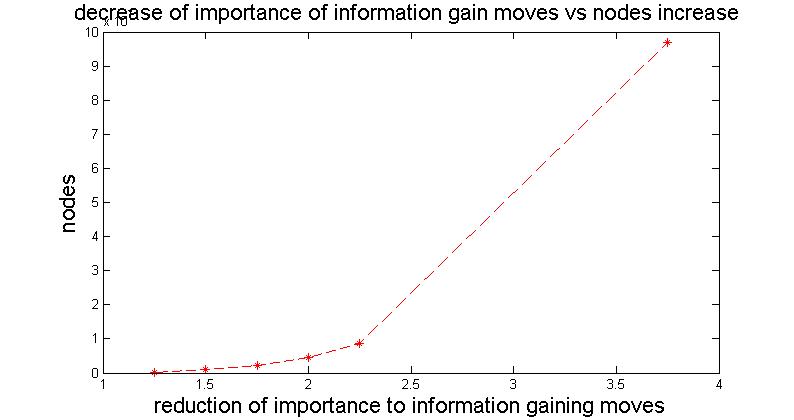} \newline
	
	The plot of nodes searched vs depth when the solution is detected fast shows a far less pronounced combinatorial explosion then when the solution is not found.          The plot shows the explosion of nodes required to find a solution when the importance given to high information lines is decreased. As the importance given to 
high information lines is decreased the number of nodes searched has to be increased.
           It increases even faster when the decisive line is not detected. For a high depth of search, the search cost registers an explosion when no decisive move is found
	reasonably fast.

	When less importance is given to high information gain moves the number of plies has to be increased to compensate this and the number of nodes explodes with the
	number of plies.
	The plot shows the necessary increase of depth when the importance of high information gain moves is decreased.

	For the case when the problem is solved the plot is: \newline

	\includegraphics[width = 2.5in , height = 2.5in]{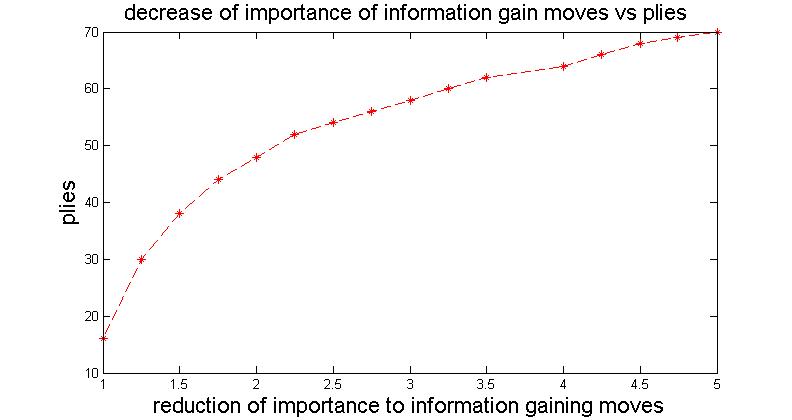} \newline

	Now we can analyze the data for the cases when the solution is not achieved. For the case when the position is not solved is a similar plot but the search at the respective depth has been realized at a far greater cost than when the solution has 			been found fast: \newline

	\includegraphics[width = 2.5in , height = 2.5in]{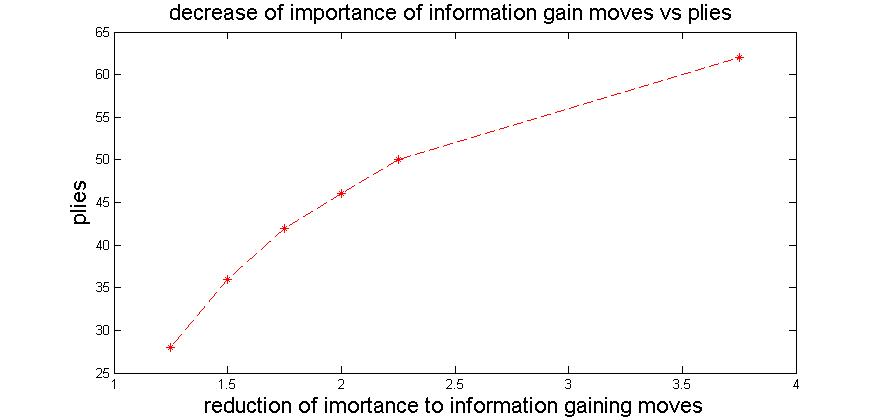} \newline

	\subsubsection{   (\romannumeral 4) The maximum depth and the importance given to high information lines }

	The maximum depth achieved decreases with a decrease in the importance given to areas of the tree with high information.
	Maximum depth vs importance given to information gain. If less importance is given to moves with high information gain more resources are needed for attaining a maximum given 		depth. This is the case for solving some combinations.

	\includegraphics[width = 2.5in , height = 2.5in]{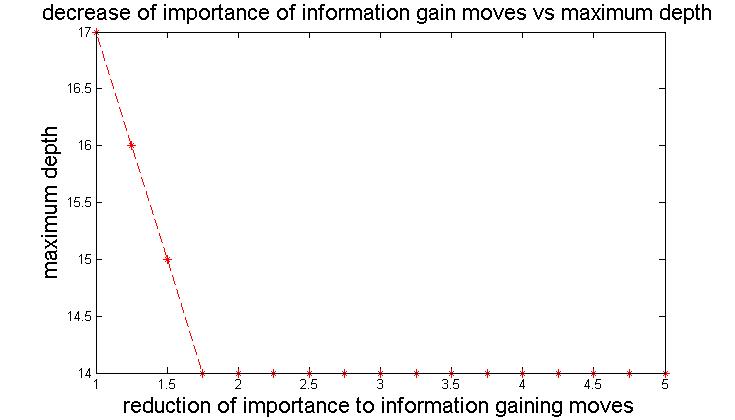} \newline

	As it can be seen from the previous plot, the maximal depth has been achieved also when the solution has not been found but as it can be observed from the above table and plots, at an ever increased cost.

	For the search experiments when the solution has not been found the highest depth remains the same but this time the cost of resources needed to sustain that depth increased very fast, faster than in the previous plot when the solution has been found.

	\includegraphics[width = 2.5in , height = 2.5in]{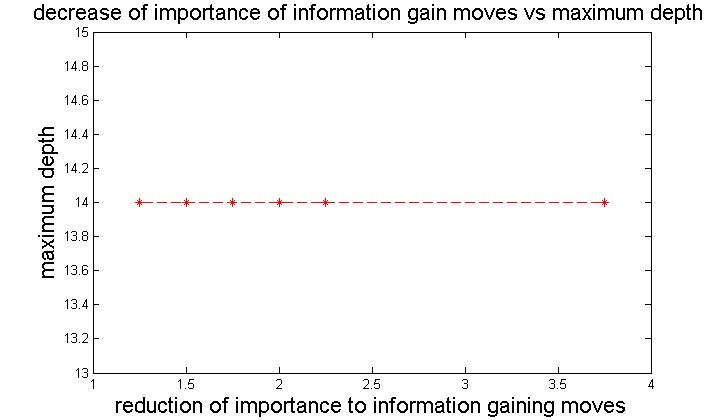} \newline

	\subsubsection{   (\romannumeral 5)  The depth of search and the objective }
	The search detects the mate even if the maximum length is just one ply deeper than the length of the combination.
	Even if we keep the maximum depth constant at far greater cost the searches are less likely to find the decisive lines as it can be seen from plots. 

	\subsubsection{  (\romannumeral 6) The effect of partial depth on the distribution of depths }

	As the search has been changed and less importance has been given to interesting moves, the range in the length of the variations became smaller as less energy
has been allocated for the most informative search lines than previously and more energy to the less informative lines. After shifting ever more resources from 
the informative line to other lines, the objective, the solution of the combination, has not been attained any more by the best lines who did not have the energy this time to penetrate deep enough. The best variations did not have any more the critical energy to penetrate the depth of the state space and solve the problem. The weaker lines
were not feasible as a path for finding any acceptable solution. From this we can understand the fundamental effect of resource allocation. And how marginal shifts in resources
can lead in this context to completely different result. If somebody used the same depth increase for each move, therefore allocating the resources uniformly to the 
variations only a supercomputer can go as deep as it is needed for finding the solution to this combination which is not among the deepest. With the introduction of 
knowledge and heuristics much greater performance would be possible. The experiment concentrates on one heuristic and its effect on the search is highly significant.

	\subsubsection { (\romannumeral 7) On deep combinations  }

	In order to solve deep combinations where some responses are not forced  a program must have chess specialized knowledge 
( or an extension of the information theoretical model of computer chess to all chess theory ) in order to give importance to variations without active moves but with significant tactical maneuvering between forceful moves such as checks and captures.

	\subsection{ Consequence 16: A theoretical justification of the formula obtained in  ~\cite{MWinands} }

	The formula obtained in this paper from information theory considerations also has e very good experimental verification in the
	formula obtained by ~\cite{MWinands} and their results.
 
	\begin{equation}    FP = \frac{\log P_c}{ \log C }   \label{EQ}\end{equation} 

	where $P_c$ is obtained empirically by observing how many moves in that category have been played from all the 	circumstances when it has been possible to play that move.

	The formula of ~\cite{MWinands} is to some extent handcrafted as are most of the formulas for calculating plies or other elements custom made 
	for games. 

	However this formula may be obtained from the formula used for the algorithm developed in out paper under some intuitive assumptions.
	In addition it is shown how to obtain the constant used by Winands in his formula. Like the formula of Winands the formula obtained
	here is a general one useful also for other games than chess.	

	Starting from 

	\begin{equation}   D_i = 1 - \frac{\triangle Entropy_{category_i} } { \triangle Entropy_{BestGain} }   \label{EQ}\end{equation} 

	Consider a capture or other operation which results in a difference in entropy of  $\log M$ and consider the best reduction of entropy achieved by 
	a move having the form $\log C$. 

	Then the following formula is obtained

	\begin{equation}   D = 1 - \frac{\log M } { \log C }   \label{EQ}\end{equation}

	This is equivalent to

	\begin{equation}   D = \frac{\log C } { \log C } - \frac{\log M } { \log C }   \label{EQ}\end{equation} 
	
	then

	\begin{equation}   D = \frac{ \log C   -   \log M } { \log C }   \label{EQ}\end{equation} 

	\begin{equation}   D = \frac{   \log \frac{ C } {M}  } { \log C }   \label{EQ}\end{equation} 
	
	then

	\begin{equation}   D = \frac{   \log { (  \frac{M}{C} ) }^{-1}   } { \log C }   \label{EQ}\end{equation}

	taking out the minus in front 

	\begin{equation}   D = (-1) \frac{   \log \frac {M}{C}   } { \log C }   \label{EQ}\end{equation} 	

	then putting the -1 at the denominator, 

	\begin{equation}   D = \frac{   \log \frac {M}{C}   } {  \log \frac{1}{C}  }   \label{EQ}\end{equation}

	therefore

	\begin{equation}   D = \frac{   \log \frac {M}{C}   } { C_2 }   \label{EQ}\end{equation}

	It can be seen in the research of ~\cite{Tsuruoka} that the captures of pieces care more often executed if possible when the piece that can be captured is 
	of high value. The value of the piece correlates with the mobility as it can be seen from the first plot and therefore there will be a significant correlation between the 			empirical probability of a category obtained by ~\cite{Tsuruoka} and the fraction $\frac {M}{C}$. Therefore the formula of Winands will also correlate to a significant 			extent with the formula obtained by using the information theoretic model.

	\subsection{ Consequence 17: A theoretical justification of the formula given by ~\cite{Tsuruoka}}

	The paper   ~\cite{Tsuruoka} and its empirical results refer to Shoji which is a form of chess popular in Asia but there are of course differences
	from the game of chess as it has been standardized by the chess organizations.	
	The empirical factor $P_c$ is very much related in many games including Chess and  Shoji with  both the weight of the piece and the mobility of the piece.

	\begin{equation}  P_c = \frac{n_p}{n_c}   \label{EQ}\end{equation}

	If the partial depth is given by the above formula then the depth factor added with each level is similar to that of the method of ~\cite{MWinands}.
	As it has been discussed above $P_c$ as calculated by ~\cite{Tsuruoka} correlates with $\frac {M}{C}$  from the information theoretical model of chess discussed here.

	Because the decisions to continue the search based on the probability factors $P_c$ generate variations having associated the probability $P_{c_1}*P_{c_2}*....$ then the partial depth factors are of the form given by the sum of logarithms of the probabilities.  As outlined above, this probability model can be seen as a way to correlate the decisions on moves to moves and strategies from knowledge-bases.

	\subsection{ Consequence 18:  A theoretical justification of the formula presented  by  ~\cite{SEX} in the selective extension algorithm 
	based on the new model. }

	The probabilities in the previous formula are determined by how likely
	such move is on the principal variation. This means how likely is the current path to be a principal variation. The answer is determined to a high extent by the type of 			moves involved.
	
	If we add the logarithms resulted from the entropy reduction at each step in depth, then we arrive at something similar to the description of the
	selective extension method.

	It is possible to obtain the formula used for the selective extensive algorithm ~\cite{SEX} 

	\begin{equation}      log[  P( M_i ) ] + log[ P( M_{ij}  ] + log[ P( M_{ijk} )  ]     \label{EQ}\end{equation}
 
	from the information theoretic model by observing that probabilities involved can be seen in the light of the information theoretic model as a 
	sum of terms of the form 
	
	\begin{equation}  \sum {D_i}  = \sum {   \frac{   \log \frac {M_{i}}{C}   } {C_2 }  }   \label{EQ}\end{equation}

	If we consider the fraction $\frac {M}{C}$ as a measure of the probability that a move is on the principal variation then , the formula obtained 			previously  explains the formula and the principle used for the selective extension search ~\cite{SEX} .

	This can be understood by observing that the moves responsible for the highest entropy reduction are also the moves very likely on the principal 
	path especially in combinations. 
 
  	As it has been shown in the description of the model, the semantic uncertainty explains why moves with big reduction in the entropy of the 
	variables modeling the position and the values are more likely to be on the principal variation. 

	For example the lines with checks only are a small percent of all variations but account for a significant number of mates.Therefore the semantic value of a string with checks is very powerful in decreasing the uncertainty and therefore the entropy in respect to the possibility that the search line may be on the principal variation.

	\subsection{Consequence 19: The endgame tables and the compression problem  }

	Information theory gives certain limits in compressing data. This may be seen also in computer chess, where endgame table-bases store 
	certain classes of end-games. Here it is very clear the relation between the entropy of the game and the size of the code needed to give 
	a perfect decision.  

	Here is the link between the problem of decreasing the entropy on a position after search and the problem of minimizing the space required for the end-game table. 
  
           \begin{conjecture} 

		Let the entropy of a position for an optimal search heuristic in chess be $H_x$ and the optimally compressed endgame table-base capable returning the perfect result for any legal and reachable position be L. Then the relation holds:

	\begin{equation} H_x \leq  L \leq H_x + 1  \label{EQ}\end{equation}

	\end{conjecture}

	In particular for the initial position, the algorithmic complexity of the game is given by the size of a database that optimally stores all the data needed for a fast and optimal first move.

	This is very intuitive because there is a relation between the uncertainty on the value of a position and the size of the database needed to store a 
	perfect result for it.There are endgame table-bases for few pieces having a size for which is sufficient the  . Practically the relation between search and retrieval from database in chess is analog to the relation between codding and decoding in 
	information theory. The calculation effort will be proportional to the uncertainty. Often the size requirements for database end-games are 				correlated to the amount of calculation needed to obtain the perfect value by search even in the case of imperfect evaluation functions. It is a 
	conjecture because it fits the present experiments in search without having a proof or having an optimal function to experiment with.

	Evidence for the connection between the number of pieces and the uncertainty consist in the increase of database size for additional endgame pieces. The endgame 
	table-base is basically an instantaneous code that gives the optimal result for a certain position. While the end-game table-bases may not be optimally compressed,
	there is significant evidence that additional pieces increase the size of such database and therefore pieces represent uncertainty in computer chess. 
It is beyond coincidence that advanced endgame positions that 
	require more calculations need a higher storage capacity for providing the perfect solution.

	A channel can be interpreted as an abstraction of a classical device including an evaluation function taking the perfect value of a position and returning an evaluation of that position, which can be modeled as the output of the channel affected by noise. In this way an evaluation function 
implemented as software or hardware can be seen as a  communication channel under noise noise.  	
Also in the past evaluation functions in chess have been simulated as a perfect evaluation function under noise by Ivan Bratko at all in 
~\cite{IB1} ~\cite{IB2} ~\cite{IB3} ~\cite{IB4} ~\cite{IB5}   and Dana Nau ~\cite{DSNau1} ~\cite{DSNau2} ~\cite{DSNau3} ~\cite{DSNau4} ~\cite{DSNau5} but they did not model this as a noisy channel. They added the noise to a perfect function in order to study the possible pathology of 
search in game trees. Channels may be interpreted as a generalization of the evaluation functions used in chess programs. 

The channel capacity C is defined in terms of the mutual information $I_{x,y}$ of the variables X (the bit sent) and the received signal Y 

 \begin{equation} C = max_{P(x)} I_{x,y} = max_{p(x)} \sum_{x \in X, y \in Y} p(x,y) \log { \frac{ p(x,y) } { p(x) p(y) }   }  \label{EQ}\end{equation}

In an analogous way to the channel capacity can be defined the accuracy of the evaluation function based on the mutual information between the evaluation of a perfect evaluation function and the evaluation of a real function. For this x is the value returned by the perfect evaluation function,
y is the value returned by the real function measured , p(y) is the probability that the perfect evaluation function will return the value V considering 
the evaluation applied on all feasible positions,  p(x) is the probability that the measured function will return the value V over all feasible positions in the
state space and p(x,y) is the probability that both will return the same value V for a position. 

 \begin{equation} A = max_{P(x)} I_{x,y} = max_{p(x)} \sum_{x \in X, y \in Y} p(x,y) \log { \frac{ p(x,y) } { p(x) p(y) }   }  \label{EQ}\end{equation}

	A search procedure with a better evaluation function would fundamentally be able to extract more information from the exploration of the state space than a search
procedure with less ability to extract information.
This information is used to guide the search and is also a result of search, suggesting a systemic relation.

	\section{Discussion}

	\subsection{General considerations}

	\paragraph{Stochastic modeling in computer chess}

	In the context of game theory, chess is a deterministic game. The practical side of decision in chess and computer chess has many probabilistic elements. The decision is deterministic, but the system that takes the decision is not deterministic, it is a stochastic system. The human decision-making
system and its features such as perception and brain processes are known to be stochastic systems. In the case of computer chess many of the search processes are also stochastic, as it has been seen from the previous examples. Considering the values of  positions in a search in depth, it is not possible to predict the value of the positions to far or even to the next such level of search
without actually searching the level. Therefore, we may consider this search as similar to a stochastic process or even to a random walk. The moves chosen finally will not be random
for a good player or a good chess program, but the values of positions obtained during the exploration of variations searched are to a significant extent random. It is a significant random factor in the estimation of the quality of a variation before it had actually been searched. A stochastic description of this type is the essence of the 
model proposed. This would be closest to the reality of decision-making processes in man and in machine. The methods of computer chess are actually probabilistic heuristics, searching a small part of the state space of chess and the performance of these procedures is probabilistic. Contrary to the
expectation of people outside the field, computer chess depends very much on probabilistic elements, even if these probabilistic elements have not been quantified to much. A general version of the problem, the NxN chess
will depend even more on probabilistic elements.These probabilistic elements can be the object of a stochastic and information theoretic approach to decision in strategic games where the state space is to big to be explored completely.  Among the probabilistic elements in computer chess are the following:
(\romannumeral 1) Having a statistical element in the machine to make the decisions of the program less predictable. This has actually proposed by Shannon.~\cite{Shannon} 
(\romannumeral 2) The idea of the selective extension algorithm, the selection of a node for search is determined probabilistically as it has been seen in several articles described previously.
(\romannumeral 3) The B* probability based search is actually named as a probabilistic method.      ~\cite{HBerliner2}                    
(\romannumeral 4) The pathology of game trees is based on a probabilistic analysis of game trees by D. Nau. 
~\cite{DSNau1} ~\cite{DSNau2} ~\cite{DSNau3} ~\cite{DSNau4} ~\cite{DSNau5}
(\romannumeral 5) Decision at strategical level in regard to choosing a strategy based on a certain risk profile. This appears as an example in Dynamic programming and optimal
control, by D. Bertsekas.

	The use of variables  which are in essence random, in the construction of systems capable of taking 
decision in strategic games justifies the stochastic analysis of these decision systems. In essence the approach may be described as a stochastic and information theoretic analysis of decision in strategic game playing systems. If one considers the NxN board a representation of the world map, a military system based on objectives and trajectories 
in both defense and attack would have certain similarities to the problem analyzed. These similarities would be about the correlations in the fields represented by possible
trajectories, high probability paths, decreasing uncertainty by technical and human intelligence methods.

	\subsection{The scope of the results}

	The information theoretic model of search presented here attempts a new perspective on computer chess and in general in game tree search. 
          Since information is important
	in both exploration and decision in computer chess, it may be possible the extension of the model to a wider range of 
	algorithms , heuristics and concepts from computer chess and strategy games in general.

	\subsection{The limitations of the model}

	The model constructs a theoretical basis for computer chess, however some of the practical methods used over many
	years in computer chess may remain outside the model. It is important that core concepts as well as some of the most powerful methods of computer chess and other 			strategy games can be described using information theoretic concepts. 

	\subsection{outlook}

	It can be expected that a significant part of the methods used already in computer chess and other strategy games can be explained using the concept 
	of information as understood in information theory. If moves such as captures, checks and so on result in high gain in 
	information, then it means heuristics such as quiescent search and search extensions follow the lines of highest information 
	gain. It may be possible to describe the entire field of computer chess and a significant number of known results using 
	concepts of information theory. Future work may show that a more extended model can be realized. The article has already
	found a number of  important results in computer chess that can be explained through information theory, confirming the 		validity of the information theoretic model in computer chess as developed here. It may be conjectured information theory can give us a precise quantification 
	of the structure of evaluation functions and other elements of systems used in decision making in chess and other strategy games.

	\subsection{Conclusion}

\paragraph{VERIFICATION OF THE THEORY DEVELOPED}

The theory, based on using information as the most important element in controlling search in strategy games, has a very strong basis:

(\romannumeral 1) The mathematical concepts used to model the theory and the relations are derived from the axiomatic foundation of information theory  
Concepts such as information gain in computer chess and other games, optimality of move ordering , the information side of chess combinations ,  the information side of tactical and strategic patterns and the architectural representation in most programs  ,  
the information theoretical meaning of quiescence search, the implication of information for pathology in game trees,
the end-game table-base,the end-game knowledge base,  have been modeled and fitted the theoretical description.

(\romannumeral 2) Some of the most important ideas are shown as a result of logical consideration offering a second verification
for the mathematical layer \newline

(\romannumeral 3) The model as seen from chess is unquestionable reflecting the domain very well. Actually the model is itself the 
formalization of the descriptions of decision in chess as offered by famous player. 

(\romannumeral 4)On the economics side, information is a very important element of decision in general, so the general idea of the 
model is correct also in economics, but of course the information theoretical model developed by in this paper quantifies information in chess.

(\romannumeral 5)The cause effect relation is very strong, no less than 20 important results or heuristics from computer chess and other games can be seen as consequences and sometimes can be derived directly from the model.

(\romannumeral 6) The experiments undertaken for the papers are novel and showing the quality of the formula and of the algorithm
developed using the formula and support even more the model.

The model starts from the axiomatic framework of information theory and describes in a formal way the role of information in 
the efficiency and effectiveness of the heuristics used in computer chess and other strategy board games.

	The model proposed considers information in its formal information theoretical meaning as the objective of exploration and the essential factor in the quality of decision in chess and computer chess as well as in other similar games.

 The method of partial depths scheme, well known in practice has
	been described mathematically by observing the fundamental fact that information gain is the criteria that determines the decrease in the uncertainty of the position. The uncertainty of the position is described in a mathematical way through the concept of entropy.

	The information gain  describes in a information theoretic way the decrease in uncertainty resulted from making a move. In this way, a quantification of search information is realized.

	This refers to entropy as it is understood in information theory but it is possible to build parallels also with thermodynamics.

	Previous approaches relied on intuitive formulas and descriptions of the best moves in terms of ''interestingness'' or in terms of chess theory or using knowledge extracted from the games of strong players.

	The approach of the method proposed here is different in that it explains an important method such as the fraction ply method using mathematical methods and formulas that can be derived from the axioms of information theory and determines important coefficients such as the fraction ply associated with moves. The method proposed here
generalizes previous approaches and grounds them on a strong theoretical axiomatic system.   

The corroboration of a significant amount of facts such as the very good integration of information theory concepts in the theory of chess and the derivation of a significant number of results and facts well known in computer chess as consequences of the information theoretical model corroborates to 
show the correctness of the description. This can be generalized to other games sharing similarities with chess as well as to the NxN chess problem. 
A better understanding of the optimization methods for EXPTIME problems is a distant and greater objective of the approach.  

A program has been implemented for showing how the calculated formula works. It contains a simple alpha-beta procedure implementing the partial depth scheme. 
In order to isolate the performance of the procedure, no chess specific knowledge is added, and a simple evaluation function using only the material value is used. This does not mean the use of chess knowledge since most of the values of the pieces can be derived from their mobility .No other heuristic is used in order to enhance the performance of the procedure proposed, because this would defy the purpose and enhance the result of the formula and
procedure presented here. Using simple heuristics and some knowledge the search could be oriented easily to target objectives more effectively in combinatorial positions. There is no element in the procedure used for search experiments to enhance the performance in this specific case. In this conditions the procedure has been often able to succeed in combinations as deep as 13 moves containing surprising moves such as leaving the queen to be captured and other moves. For this case and many like this the program found the solution searching somewhere between 10 thousands and one million moves, which is very few. For comparison, a brute force alpha-beta would often require more than a billion moves for searching uniformly 14 plies in depth. Only a very powerful supercomputer could execute a brute force without extensions and quiescence and other heuristics over 14 plies in depth in the state space of chess. Of course, top programs use many heuristics, not just brute force alpha-beta with constant size for the plies. 

 Extension heuristics are explained in the context of partial depth scheme. A new description of evaluation functions for chess , mutual information of positions, information gain in chess search and other elements are presented in the information theoretical framework.

	Evidence by testing a program with test cases is provided. A significant body of experimental knowledge from computer chess and the tests performed by all those who used the fractional ply method and various formulas of this type chosen intuitively or handcrafted or by experience provides the very strong experimental verification of the model. 

A number of results existing before in computer chess have been justified through the model. Among these, but not only these are the following:
the endgame tables, a theoretical justification of the formula presented  by  ~\cite{SEX}, of the formula presented by ~\cite{MWinands}  ,  of the formula presented by  ~\cite{Tsuruoka} . A foundation for one of the best methods of search in computer chess, the selective extension method is given using the information theoretic model. A new formula is derived here from first principles and tested through an implementation. It is shown how the fraction ply method generalizes the brute force alpha-beta, by giving more importance to moves and lines providing high information in the sense measured in the model. The model describes a mathematical representation of evaluation functions and opens the way to mathematical model of the 
chess and strategy programs with a quantitative view on the structure, something not yet done. A reason explaining why some heuristics such as quiescent search work is provided: they filter moves selecting transitions with high information gain and transitions towards low entropy and equilibrium positions.  

 It can be expected that information, in the sense of information theoretic concept is the key to modeling both decision and exploration in computer chess and in many other strategy games. Information plays an important role in decision in general. 

Application of such information driven methods are exploring the web in the areas of greatest information for commercial use and
for military use (adversarial search)  the modeling of intelligence operation driven by goals and allocating the resources on the
lines of highest information gain, military strategy based on the concept of field of trajectories as it could be derived if one transforms
a NxN board in a world map with meridians and parallels. The goal is to reduce the uncertainty on the highest probability trajectories
and increase the uncertainty in the adversarial model of search.

 The conclusion is that a stochastic model of search is closer to the way humans think chess and how computers decide on chess moves. The quantification of the stochastic elements of search may be the key to deeper understanding of game decision and of decision in general.	It may be hoped this model is powerful enough to upgrade the computer chess model of Claude Shannon obtaining new advancements.  Similar models may be used for other search problems, games , strategy processes and for a deeper understanding of search in general.

	\bibliographystyle{plain}

\begin{thebibliography}{100}
		\bibitem{MarshlandT} T.A. Marshland, computer chess methods
		\bibitem{SchaefferJ} Johnathan Schaeffer, The history heuristic and alpha-beta search enhancements in practice
		\bibitem{Shannon} Claude Shannon, Programming a computer for playing chess
		\bibitem{StrongG} Glenn Strong, The minimax algorithm 
		\bibitem{MarshalCampbel} T.A. Marshland, M. Campbell, Parallel Search of Strongly Ordered Game Trees
		\bibitem{RFine} R.F. Chess the easy way
		\bibitem{Kotov} Kotov, Think like a grandmaster
		\bibitem{wikiG}  the wikipedia page of Games Theory as of November 2011 
		\bibitem{wikiCC} the wikipedia page of computer chess as of November 2011
		\bibitem{wikiCGO} the wikipedia page of computer GO as of November 2011
		\bibitem{wikiCG} the wikipedia page of complexity of games as of November 2011
		\bibitem{wikiE} the wikipedia page of thermodynamic entropy as of November 2011
		\bibitem{wikiIT} the wikipedia page of information theory as of November 2011
		\bibitem{wikiTE} the wikipedia page of thermodynamic entropy as of November 2011
		\bibitem{wikiTEITE} the wikipedia page of thermodynamic entropy and IT entropy as of November 2011
		\bibitem{AGTesis} Alexandru Godescu, Thesis for the degree of Engineer in computer science , PUB 2006
		\bibitem{AGCS} Alexandru Godescu, Case Study presentation ETH Zurich 2011
		\bibitem{GTOR} Martin J. Osborne, Ariel Rubinstein, A course in game theory 
		\bibitem{TOGEB} Theory of games and Economic behavior, Commemorative Edition, (Princeton Classic Edition) John von 						      Neumann, Oskar Morgenstern, Harold William Kuhn and Ariel Rubinstein
		\bibitem{FL} S. Fraenkel and D. Lichtenstein, Computing a perfect strategy for n*n chess requires time exponential in n, 				         Proc. 8th Int. Coll. Automata, Languages, and Programming, Springer LNCS 115 (1981) 278-293 and J. Comb. 				         Th. A 31 (1981) 199-214
		\bibitem{DSNau1} D.S. Nau, Quality of decision versus depth of search on game trees. In: (2nd ext. ed. ), Ph.D. Thesis, 						Duke University, Durham, NC (1979).
		\bibitem{DSNau2} D.S. Nau, Pathology on game trees: a summary of results. In: Proceedings AAAI-80 (1980), pp. 				102-104.
		\bibitem{DSNau3} D.S. Nau, An investigation of the causes of pathology in games. Artificial Intelligence 19 3 (1980), pp. 			257-278
		\bibitem{DSNau4} D.S. Nau, Decision quality as a function of search depth in decision trees, J. ACM 30 4 (1983), pp. 				687-708
		\bibitem{DSNau5} D.S. Nau, Pathology on game trees revisited, and an alternative to mini-maxing. Artificial Intelligence 21 			1-2 (1983), pp. 687-708
		\bibitem{Botvinnik1} M.M. Botvinnik, Computers in chess: solving inexact search problems
		\bibitem{Botvinnik2} M.M. Botvinnik, Computers, chess and long range planning
		\bibitem{HBerliner1} H. J. Berliner, A chronology of computer chess and its literature
		\bibitem{HBerliner2} H. J. Berliner, The B* tree search algorithm: A best-first proof procedure, Artificial Intelligence 1978 
		\bibitem{RK} R. Keeny, The Chess Combination from Philidor to Karpov, Learn tactics from champions
		\bibitem{BB} B. Br\"ugmann, Monte Carlo GO , Max-Plank-Institute of Physics
		\bibitem{Markowitz} H.M. Markovitz (March 1952) "Portfolio Selection". The Journal of Finance 7 (1): 77-91
		\bibitem{CoverThomas} T. Cover, J. Thomas, Elements of Information Theory, Second Edition
		\bibitem{MezardMontanari} Marc Mezard, Andrea Montanari , Information, Physics, and Computation   
		\bibitem{PijlsBruin} Wim Pijls, Arie de Bruin, Searching Informed game trees
		\bibitem{EliotSlater} Eliot Slater,  Statistics for the Chess computer and the factor of mobility 
		\bibitem{RISK} F. Etiene De Vylder, Advanced risk theory, a self contained introduction
		\bibitem{Knuth} D E Knuth  An analysis of Alpha-Beta Pruning,Artificial Intelligence 1975 
		\bibitem{Caissa}G.M. Adelson-Velsky, V.L. Arlazarov,M.V. Donskoy,  Some Methods of Controlling the Tree Search in Chess 			Programs, Artificial Intelligence 1975
		\bibitem{relEntropyWiki} The wikipedia page of the relative entropy as of November 2011
		\bibitem{infoGainWiki}     The wikipedia page of information gain in decision trees as of November 2011
		\bibitem{decisionTrees}   The wikiepdia page of information gain   	
		\bibitem{Renner} Renato Renner, Lecture Notes, Quantum Information Theory, August 16, 2011
		\bibitem{JPearl} Judea Pearl, On the Nature of Pathology in Game Searching
		\bibitem{MWinands} Mark Winands, Enhanced Realization Probability Search
		\bibitem{SEX} David Levy, David Broughton, Mark Taylor (1989), The SEX algorithm in computer chess ICGA 			Journal, vol. 25, No. 3
		\bibitem{Tsuruoka} Game-Tree Search Algorithm based on Realization Probability, ICGA Journal, Vol. 25, No. 3
		\bibitem{Solomonoff} Ray Solomonoff , ''The Universal Distribution and Machine Learning'', The Kolmogorov Lecture, Feb 27, 2003, 
		Royal Holloway, Univ. of London. The Computer Journal, Vol 46, No. 6, 2003.
		\bibitem{JS} Jonathan Schaeffer, Andreas Junghanns, Search Versus Knowledge in Game-Playing programs Revisited, 
		\bibitem{IB1} Aleksander Sadikov, Ivan Bratko, Igor Konenko ,Search vs Knowledge: empirical study of minimax on KRK endgame
		\bibitem{IB2} Dana S. Nau, Mitja Lustrek, Austin Parker, Ivan Bratko, Matjaz Gams, When is better not to look ahead ?
		\bibitem{IB3} Mitja Lustrek, Matjaz Gams, Ivan Bratko, Is Real-Valued Minimax Pathological ?
		\bibitem{IB4} Mitja Lustrek, Matjaz Gams, Ivan Bratko, Why Minimax Works: An Alternative Explanation
		\bibitem{IB5}  Aleksander Sadikov, Ivan Bratko, Igor Konenko, Bias and pathology in minimax search
	\end{thebibliography}

          \section{APPENDIX}

	\appendix

	\section{Code}

\begin{verbatim}
	double minimax(double alfa,double beta,int depth,int k,int type,move mv,
							double previousval,double virtualdepth){
	move* listNewMoves = (move*) new move [100];
	move mr;    double value = 0 , temp = 0 , ev = 0 ;  int c,number;

	if( (virtualdepth >= maxDepth || depth >= maxExtension ) ){		
		 return evaluation(type,mv); 
	}else{
		if( tip == 1 ){     value = -10000;         }
		else{
			value = 10000;
		}
	
	generator(mv,listNewMoves,number);
	
	for(int i=1; i <= number ;i++){
		listNewMoves[i].eval = fabs( evaluation(tip,listNewMoves[i]) -  previousval ) ;
		double b = -1;
		if( isCheck( listNewMoves[i] ) )
			listNewMoves[i].eval += 10000;	
	}
		
    if( number == 0 ){
		if( tip == 1 )
			if( !is_legal_w(mv) ) return inf_plus;
			else return 0; 
		}else{
			if( !is_legal_n(mv) ) return inf_neg;
			else return 0;
		}
	}else
		for(int k1=1;k1 <= number;k1++){	
			double max = -1;
		int ic = -1;
		for(int c = 1 ; c <= number  ; c++ ){
			double comp = listNewMoves[c].eval;
			if( comp > max ){
				max = listNewMoves[c].eval; 
				ic = c;
			}
		}
	
		mr.eval = listNewMoves[ic].eval;
		double evalPosition = listNewMoves[ic].eval;
		lista_pozitii_urm[ic].eval = -2;
		copy(mr.move, listNewMoves.move);
		copy( mr.tabla,  listNewMove[ic].tabla );
		mr.turn = lista_pozitii_urm[ic].turn;
		double nextV = evaluation(tip,mr);
	
	if(  evalPosition > 2000    )
		value = - minimax( -beta ,-alfa, depth + 1 , ic ,-tip,mr,nextV,virtualdepth );
	else {
			double add = log(fabs(0.1 + (evalPosition/100)))/(log(10.0)) + 5.0/log(number + 2 );		
			value = - minimax( -beta ,-alfa, depth + 1 , ic ,-tip,mr,nextV,virtualdepth + 6 - add);
	}
			if( value >= alfa )   alfa = value;

			if( alfa >= beta ){
				cutoff++;
				break;
			}
	}
	return alfa;
	}
}

\end{verbatim}

	\section{Contents} 

	\tableofcontents
	
\end{document}